\documentclass{article}

\usepackage{microtype}
\usepackage{graphicx}
\usepackage{subcaption}
\usepackage{booktabs} % for professional tables
\usepackage{multirow}
\usepackage[table]{xcolor}
\usepackage{array}
\usepackage{makecell}

\usepackage{hyperref}

\usepackage[accepted]{icml2026}

\usepackage{amsmath}
\usepackage{amssymb}
\usepackage{mathtools}
\usepackage{amsthm}

\usepackage[capitalize,noabbrev]{cleveref}

\theoremstyle{plain}
\newtheorem{theorem}{Theorem}[section]

\newtheorem{lemma}[theorem]{Lemma}

\theoremstyle{definition}

\newtheorem{assumption}[theorem]{Assumption}
\theoremstyle{remark}

\usepackage[textsize=tiny]{todonotes}

\icmltitlerunning{Kuramoto Oscillatory Phase Encoding}

\begin{document}

\twocolumn[
  %\icmltitle{Submission and Formatting Instructions for \\
  %  International Conference on Machine Learning (ICML 2026)}
  \icmltitle{Kuramoto Oscillatory Phase Encoding: Neuro-inspired Synchronization for Improved Learning Efficiency}



  \icmlsetsymbol{equal}{*}

  \begin{icmlauthorlist}
    \icmlauthor{Mingqing Xiao}{msra}
    \icmlauthor{Yansen Wang}{msra}
    \icmlauthor{Dongqi Han}{msra}
    \icmlauthor{Caihua Shan}{msra}
    \icmlauthor{Dongsheng Li}{msra}
  \end{icmlauthorlist}

  \icmlaffiliation{msra}{Microsoft Research Asia}%, Shanghai, China}

  \icmlcorrespondingauthor{Mingqing Xiao}{mingqingxiao@microsoft.com}

  \icmlkeywords{phase synchronization, Kuramoto model, learning efficiency, vision transformer}

  \vskip 0.3in
]



\printAffiliationsAndNotice{}  % no special notice (required even if empty)

\begin{abstract}
  Spatiotemporal neural dynamics and oscillatory synchronization are widely implicated in biological information processing and have been hypothesized to support flexible coordination such as feature binding. By contrast, most deep learning architectures represent and propagate information through activation values, neglecting the joint dynamics of rate and phase. In this work, we introduce Kuramoto oscillatory Phase Encoding (KoPE) as an additional, evolving phase state to Vision Transformers, incorporating a neuro-inspired synchronization mechanism to advance learning efficiency. We show that KoPE can improve training, parameter, and data efficiency of vision models through synchronization-enhanced structure learning. Moreover, KoPE benefits tasks requiring structured understanding, including semantic and panoptic segmentation, representation alignment with language, and few-shot abstract visual reasoning (ARC-AGI). Theoretical analysis and empirical verification further suggest that KoPE can accelerate attention concentration for learning efficiency. These results indicate that synchronization can serve as a scalable, neuro-inspired mechanism for advancing state-of-the-art neural network models. Code is avaliable at \url{https://github.com/microsoft/Neuro-inspired_Phase_Encoding}.
\end{abstract}

\section{Introduction}

Artificial neural networks are primarily built upon activation-based neurons, initially inspired by the McCulloch-Pitts neuron~\citep{mcculloch1943logical} and often treated as a rate-based analogue of biological neural activity~\citep{hopfield1984neurons}. Much of the progress in neural networks has then been driven by architectural/connectivity design, with developments from MLP~\citep{rumelhart1986learning} to LSTM~\citep{hochreiter1997long}, CNN~\citep{lecun2002gradient}, residual connections~\citep{he2016deep}, Transformer~\citep{vaswani2017attention}, etc. It is widely observed that modern neural networks can increasingly improve performance with the scaling of data, computation and model size~\citep{kaplan2020scaling,zhai2022scaling,cherti2023reproducible}, and recent progress has been strongly driven by scaling paradigms~\citep{bommasani2021opportunities,oquab2024dinov}. On the other hand, human brains typically do not rely on data at the scale used in modern large models and can often acquire new concepts, abstractions, and structured generalization from far fewer examples than current deep models~\citep{chollet2019measure,ilievski2025aligning}. This gap suggests that additional mechanisms or inductive biases from human brains for efficient learning may still be missing in mainstream neural architectures.

In contrast to activation-based neural networks, neuroscience research has long emphasized spatiotemporal neural dynamics, where travelling waves have been proposed as important mechanisms for information processing~\citep{muller2018cortical}. Beyond firing rates, oscillatory waves are characterized by their phases as a component of neural coding. A prominent phenomenon in this regime is phase synchronization for information processing, which has been implicated in coordinating neural processing across space and time~\citep{singer1999neuronal,fries2015rhythms}. 

An appealing hypothesis of synchronization is its relation to the binding problem~\citep{singer1995visual,engel2001temporal}, a long-standing problem in both neuroscience and AI related to structure learning and compositional generalization~\citep{greff2020binding}. Binding refers to the dynamic integration of distributed feature representations into coherent structured entities, such as features belonging to the same object or concept are selectively associated while remaining separable from others~\citep{treisman1996binding}. Such kind of structure enables systematic reuse and recombination of representations to facilitate generalization and data efficiency~\citep{greff2020binding}, and lacking of it was considered as a fundamental reason behind drawbacks of deep learning~\citep{greff2020binding,zheng2022dance}, especially for unstructured visual data. For phase synchronization, it was conjectured to indicate binding relations among concurrently processed features~\citep{singer1995visual,greff2020binding}, thus potentially leading to structures for generalization and efficiency. Therefore, synchronization may be a kind of inductive bias for efficient learning.

Some previous works explored synchronization in neural networks in the form of complex-valued neurons~\citep{reichert2013neuronal,lowe2022complex} or spiking time~\citep{zheng2023gust}, focusing on binding-specific problems such as object discovery. Recently, Kuramoto dynamics, a mathematical model for synchronization of oscillators~\citep{acebron2005kuramoto}, have also been introduced to formulate a phase-only neuron model that evolves for synchronization~\citep{miyatoartificial}. Yet a predominant problem for these works and other binding-like object-centric methods~\citep{locatello2020object,lowe2023rotating} is the scalability to common large-scale tasks on natural images. Most of these models are isolated from state-of-the-art scalable models, not answering how such mechanism can advance mainstream tasks. A natural question is: can phase synchronization be turned into a scalable inductive bias for modern deep learning, facilitating binding-like structure learning and improving learning efficiency without sacrificing scalability?

In this work, we introduce Kuramoto oscillatory Phase Encoding (KoPE), an additional, evolving phase state along with the mainstream Vision Transformer (ViT) model~\citep{dosovitskiy2021an}, and show that neuro-inspired synchronization can advance scalable neural networks for learning efficiency. Specifically, we incorporate phase representations for each token, which is updated across layers via Kuramoto dynamics with data-dependent coupling derived from token representations. In parallel, phases are injected into the interactive attention module through complex-form rotations, coupling phase evolution to token (rate) representations. In this way, the synchronization dynamics of phases encourage structure formation from data, providing an architecture-level inductive bias to improve learning efficiency through joint rate-phase dynamics.

We conduct extensive experiments across supervised and self-supervised learning, demonstrating that KoPE can improve training, parameter, and data efficiency. Moreover, KoPE shows superior performance for tasks demanding structured understanding, including semantic and panoptic segmentation, vision-language representation alignment, and the challenging few-shot abstract visual reasoning tasks, ARC-AGI~\citep{chollet2019measure,chollet2025arc}. 

Theoretical analysis and empirical verification further suggest that KoPE can facilitate attention learning, i.e., the structural relation among tokens, through attention concentration on relevant parts. 
Together, these results suggest a practical route for bringing synchronization-based dynamics into modern neural architectures, bridging neuro-inspired principles with scalable vision learning.

\vspace{-2mm}
\section{Preliminary}

\vspace{-1mm}
\subsection{Kuramoto Model}
\vspace{-0.5mm}

The Kuramoto model~\citep{acebron2005kuramoto} is a non-linear dynamical model describing the synchronization of coupled oscillators. It has been widely applied in neuroscience to characterize neural oscillation~\citep{breakspear2010generative}. Specifically, for each oscillator $i$, it has a phase $\phi_i\in[0,2\pi)$ that evolves following the differential equation:
\begin{equation}
    \dot{\phi_i}=\omega_i+\sum_j J_{ij}\sin(\phi_j-\phi_i),
\end{equation}

\vspace{-3.5mm}
where $\omega_i$ is the natural frequency of the oscillator and $J_{ij}$ are coupling connections between oscillators. It can also be extended to a multi-dimensional vector version~\citep{miyatoartificial}:
\vspace{-4.5mm}
\begin{equation}
    \dot{\mathbf{x}_i}=\mathbf{\Omega}_i\mathbf{x}_i+\text{Proj}_{\mathbf{x}_i}\left(\sum_j\mathbf{J}_{ij}\mathbf{x}_j\right),
\end{equation}

\vspace{-3.5mm}
where $\mathbf{x}_i\in\mathbb{R}^n, \lVert \mathbf{x}_i\rVert_2=1$ is a vector on a hypersphere, $\mathbf{\Omega}_i$ is an anti-symmetric matrix, and $\text{Proj}_{\mathbf{x}_i}(\mathbf{y})=\mathbf{y}-\langle \mathbf{y}, \mathbf{x}_i \rangle \mathbf{x}_i$ is projection to the orthogonal direction. When $\mathbf{x}_i=\left(\cos(\phi_i), \sin(\phi_i)\right)^\top$, $\mathbf{\Omega}_i=\begin{pmatrix} 0 & -\omega_i \\ \omega_i& 0\end{pmatrix}$, and $\mathbf{J}_{ij}=J_{ij}\mathbf{I}$, the vector version is equivalent to the phase version. When $J_{ij}=J_{ji}$ and $\omega_i=\omega$, the Kuramoto model is Lyapunov and minimizes the energy $U(\phi)=\sum_{i,j}J_{ij}(1-\cos(\phi_i-\phi_j))$. Oscillators will converge to a phase-locked equilibrium state with (cluster) synchronization depending on the coupling structure and initial condition.
%We will use this vector version in computation.

In practice, we discretize the differential equation as:
\vspace{-2.5mm}
\begin{equation}
    \begin{aligned}
        \Delta \mathbf{x}_i^t &= \mathbf{\Omega}_i\mathbf{x}_i^t+\text{Proj}_{\mathbf{x}_i^t}\left(\sum_j \mathbf{J}_{ij}\mathbf{x}_j^t\right),\\
        \mathbf{x}_i^{t+1} &= \frac{\mathbf{x}_i^t+\gamma\Delta\mathbf{x}_i^t}{\lVert\mathbf{x}_i^t+\gamma\Delta\mathbf{x}_i^t\rVert_2}.
    \end{aligned}
    \label{eq:kuramoto}
\end{equation}

\vspace{-2mm}
\citet{miyatoartificial} leverage the Kuramoto model to form phase-only neurons for synchronization, yet they neglect the joint dynamics of rate and phase, isolating the model from state-of-the-art neural networks. Differently, we consider a synergy of rate and phase, incorporating Kuramoto dynamics as a synchronization prior to vision transformers.

\vspace{-1mm}
\subsection{Vision Transformer}

\begin{figure*}[t]
    \centering
    \includegraphics[height=0.28\textheight]{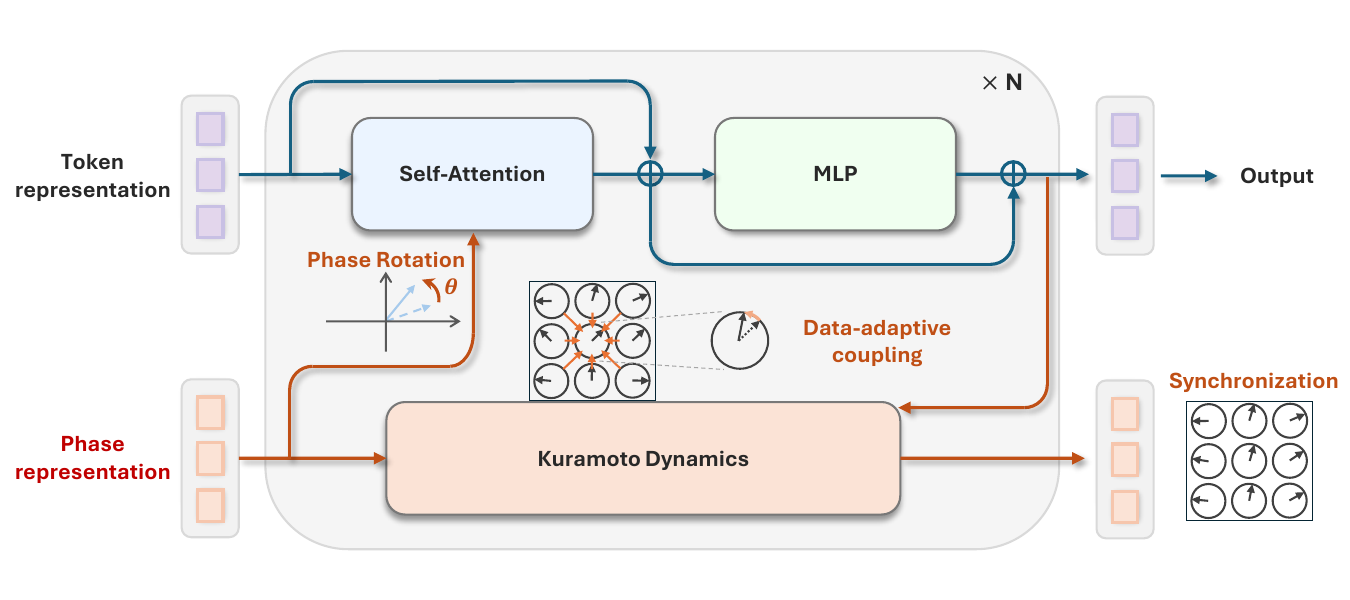}
    \vspace{-2.5mm}
    \caption{Illustration of the proposed KoPE. We introduce phase states apart from token representations, which will be updated by Kuramoto dynamics along the depth of the layers. The phases are injected into the interactive attention module through complex-form rotations, and Kuramoto dynamics are based on data-adaptive couplings derived from token representations.}
    \label{fig:illustration}
    \vspace{-2mm}
\end{figure*}

ViT patchifies images into tokens $\{\mathbf{z}_i\}$ and updates token representations through Transformer layers:
\vspace{-2mm}
\begin{equation}
    \begin{aligned}
        {\mathbf{z}^{l+1}}' &= \mathbf{z}^l + \mathrm{MHSA}\!\big(\mathrm{LN}(\mathbf{z}^l)\big), \\
        \mathbf{z}^{l+1}  &= {\mathbf{z}^{l+1}}' + \mathrm{MLP}\!\big(\mathrm{LN}({\mathbf{z}^{l+1}}')\big),
    \end{aligned}
\end{equation}
where MHSA is multi-head self-attention, MLP is multi-layer perception, and LN is layer normalization.

\textbf{Neural ODE view of ViT}\quad
In the neural ODE view~\citep{chen2018neural}, deep residual networks implement discrete time steps of an underlying continuous‑time dynamical system $\dot{\mathbf{z}}(t)=f(\mathbf{z}(t), t, \mathbf{\theta}(t))$, so their depth $l$ indexes the evolution rate of a hidden state through a learned vector field. This perspective extends naturally to Transformers (and ViTs), where the stacked updates can be interpreted as a splitting scheme for an ODE~\citep{lu2020understanding}. This dynamical system view motivates us to integrate Kuramoto dynamics of phases into the depth evolution of ViT.

\vspace{-2mm}
\section{Method}

To incorporate Kuramoto dynamics as a structural prior, we propose Kuramoto oscillatory Phase Encoding as an additional evolving phase dimension along the depth of ViT, as shown in Fig.~\ref{fig:illustration}. The phases are updated based on data-dependent attentive connections derived from token representations, with the aim of synchronizing parts related to similar concepts/entities; on the other hand, token representations are modulated by phases in the complex form within the interactive attention module, analogous to the joint dynamics of rate and phase. 

\vspace{-1mm}
\subsection{Phase Representation and Rotary Integration}\label{subsec:phase_representation}

For each token, apart from token representations $\mathbf{z}_i\in\mathbb{R}^d$, we introduce phase representations $\mathbf{r}_i\in\mathbb{R}^{2p}$ indicating the concatenation of multiple phase vectors $(\cos(\phi_{i,j}), \sin(\phi_{i,j}))$. We expect the phases to inform the relations through synchronization and phase differences, e.g., phases are synchronized for the same entity and differed for different ones. The complex form of rate and phase thus encodes both feature and relational information. Inspired by rotary position embedding~\citep{su2024roformer}, we can leverage such complex form in the attention module by rotating the query and key vectors. For a simple case with dimension $d=2$ and one phase per token, the formulation is:
%\vspace{-1mm}
\begin{equation}
    \begin{aligned}
        f_q(\mathbf{z}_m) &= (\mathbf{W}_q\mathbf{z}_m)\text{e}^{i\phi_m}, \\
        f_k(\mathbf{z}_n) &= (\mathbf{W}_k\mathbf{z}_n)\text{e}^{i\phi_n}, \\
        \langle f_q(\mathbf{z}_m), f_k(\mathbf{z}_n) \rangle &= \text{Re}\left[(\mathbf{W}_q\mathbf{z}_m)(\mathbf{W}_k\mathbf{z}_n)^*\text{e}^{i(\phi_m-\phi_n)}\right],
    \end{aligned}
\end{equation}
where the phase difference $\phi_m-\phi_n$ is naturally integrated into the dot product of query and key in the attention calculation. It can be implemented as multiplying the rotation matrix $\mathbf{R}(\phi_m)=\begin{pmatrix} \cos(\phi_m) & -\sin(\phi_m) \\ \sin(\phi_m)& \cos(\phi_m)\end{pmatrix}$ to the query/key vectors with efficient realization. In this way, phases are modulating the attention scores.

We further extend the rotation to value and output vectors so that phases modulate the value integration process as well. For the same case as above, the calculation is:
\begin{equation}
    \begin{aligned}
        f_v(\mathbf{z}_m) &= \mathbf{R}(\phi_m)(\mathbf{W}_v\mathbf{z}_m), \\
        f_o(\mathbf{z}_n) &= \mathbf{W}_o\mathbf{R}(-\phi_n)\sum_ma_{n,m}f_v(\mathbf{z}_m)
    \end{aligned}
\end{equation}

\vspace{-3.5mm}
where $a_{m,n}$ is the attention score. This enables phase interference for amplitude enhancement or attenuation. A related use of rotations on value vectors has been explored by \citet{miyato2024gta} for injecting 3D geometry in multi-view transformers, whereas our rotations are driven by evolving Kuramoto phases.

For the general high-dimensional setting where $d$ is even, we also divide the $d$-dimension space into $d/2$ subspaces~\citep{su2024roformer} and integrate phases for each one. We will take $p=d/2$ (or $p=d_h/2$ for each head in multi-head attention) to assign a phase for each subspace. 

While this complex form is inspired by rotary position embedding, we essentially differ from those methods targeted for static positions, as our objective is relational phase differences induced by endogenous Kuramoto phase dynamics, which induces structure from data through synchronization.

\subsection{Phase Dynamics}

We align the discrete temporal dynamics of phases with the depth of ViT and update them at each layer. The couplings $\mathbf{J}$ are data-adaptive attentive connections so that phases are expected to converge to cluster synchronization depending on the data. Specifically, we apply shared parameterized modules $h_q(\cdot)$ and $h_k(\cdot): \mathbb{R}^d\rightarrow\mathbb{R}^d$ over token representations at each layer and calculate coupling connections as:
%\vspace{-1mm}
\begin{equation}
    \mathbf{J}^l = \mathrm{softmax}_{\text{col}}\left(\frac{h_q(\mathbf{z}^l)h_k(\mathbf{z}^l)^\top}{\sqrt{d}}\right).
\end{equation}
Then phase representations are updated as Eq.~(\ref{eq:kuramoto}) based on coupling connections. Since we only care about the relative phase differences (Section~\ref{subsec:phase_representation}), we can omit the calculation of $\mathbf{\Omega}_i$ if all oscillators share the same natural frequency (the condition for Lyapunov).

Therefore, the calculation of each layer is summarized as:
\begin{equation}
    \begin{aligned}
        {\mathbf{z}^{l+1}}' &= \mathbf{z}^l + \mathrm{RMHSA}\!\big(\mathrm{LN}(\mathbf{z}^l), \mathbf{r}^l\big), \\
        \mathbf{z}^{l+1}  &= {\mathbf{z}^{l+1}}' + \mathrm{MLP}\!\big(\mathrm{LN}({\mathbf{z}^{l+1}}')\big), \\
        \mathbf{r}^{l+1} &= \Pi^{(2)}\left(\mathbf{r}^{l}+\gamma\text{Proj}_{\mathbf{r}^l}^{(2)}\left(\mathbf{J}^{l+1}\mathbf{r}^l\right)\right),
    \end{aligned}
\end{equation}
where RMHSA refers to multi-head self-attention with phase rotation, and $\Pi^{(2)}$ and $\text{Proj}_{\mathbf{r}^l}^{(2)}$ means pairwise (i.e., for the two elements of $\cos$ and $\sin$) normalization and orthogonal projection. The coupling also maintains the multi-head setting as in the attention module (details in Appendix~\ref{supsec:implementation_details}).

\begin{figure*}[t]
  \centering
  % ---------------- Row 1: 3 plots ----------------
  \begin{subfigure}[t]{0.32\textwidth}
    \centering
    \includegraphics[width=\linewidth]{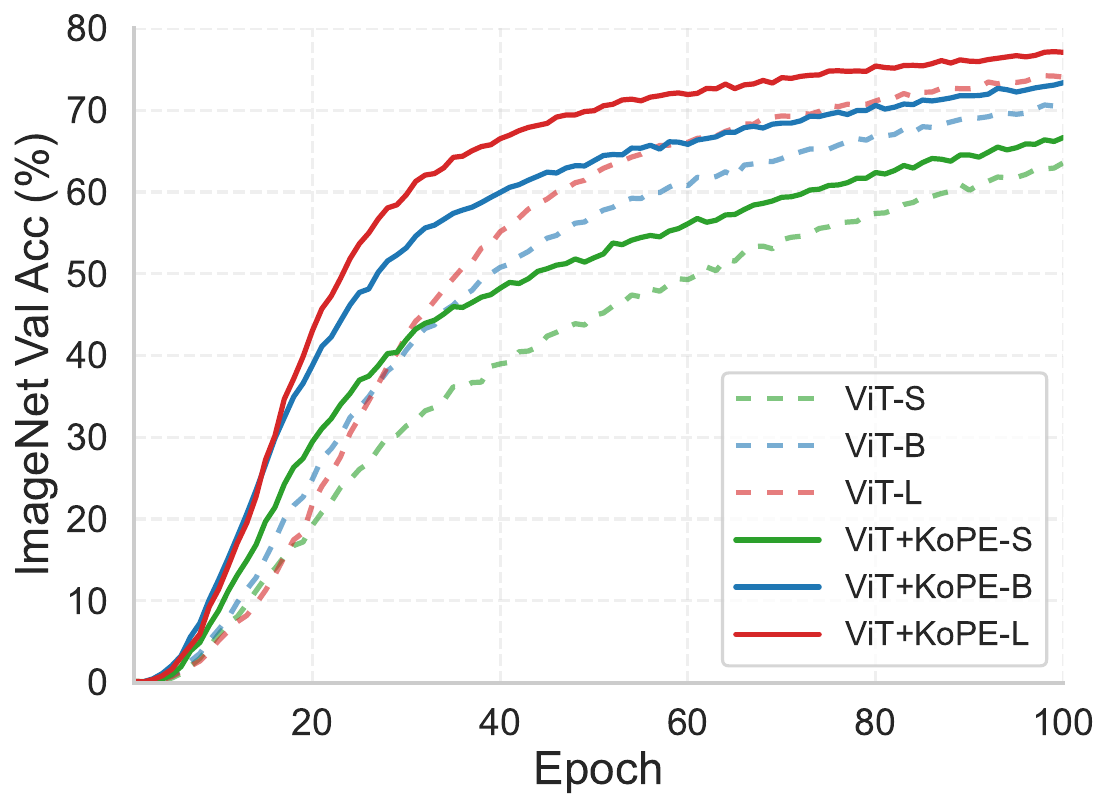}
    \subcaption{Early training stage}
    \label{fig:early-acc}
  \end{subfigure}
  \hfill
  \begin{subfigure}[t]{0.32\textwidth}
    \centering
    \includegraphics[width=\linewidth]{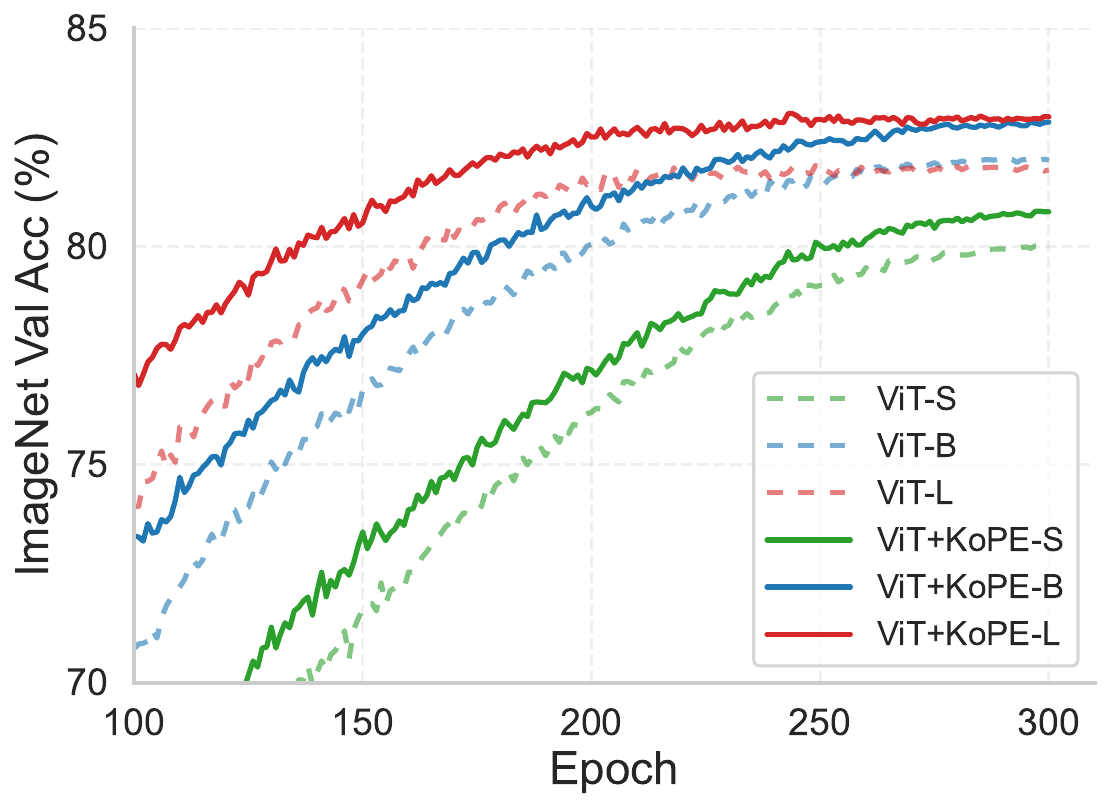}
    \subcaption{Late training stage}
    \label{fig:late-acc}
  \end{subfigure}
  \hfill
  \begin{subfigure}[t]{0.32\textwidth}
    \centering
    \includegraphics[width=\linewidth]{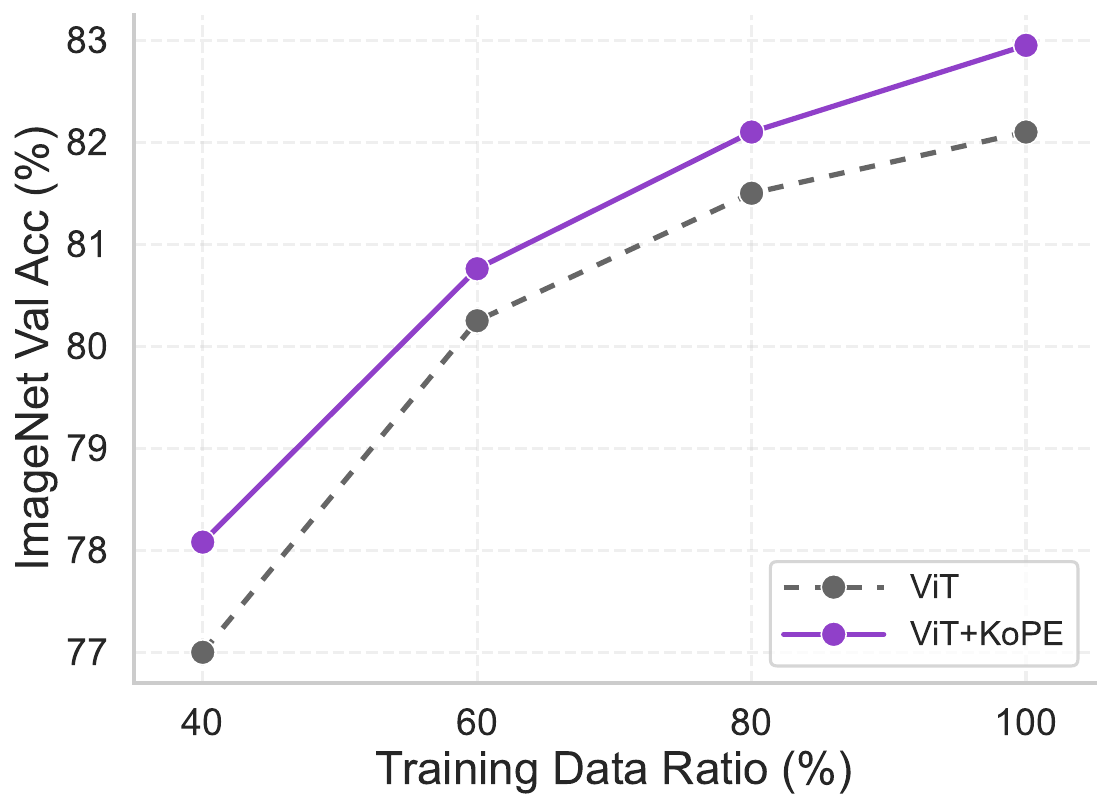}
    \subcaption{Data efficiency}
    \label{fig:data-eff}
  \end{subfigure}
  \vspace{0.6em}
  % ---------------- Row 2: 4 plots ----------------
  \begin{subfigure}[t]{0.24\textwidth}
    \centering
    \includegraphics[width=\linewidth]{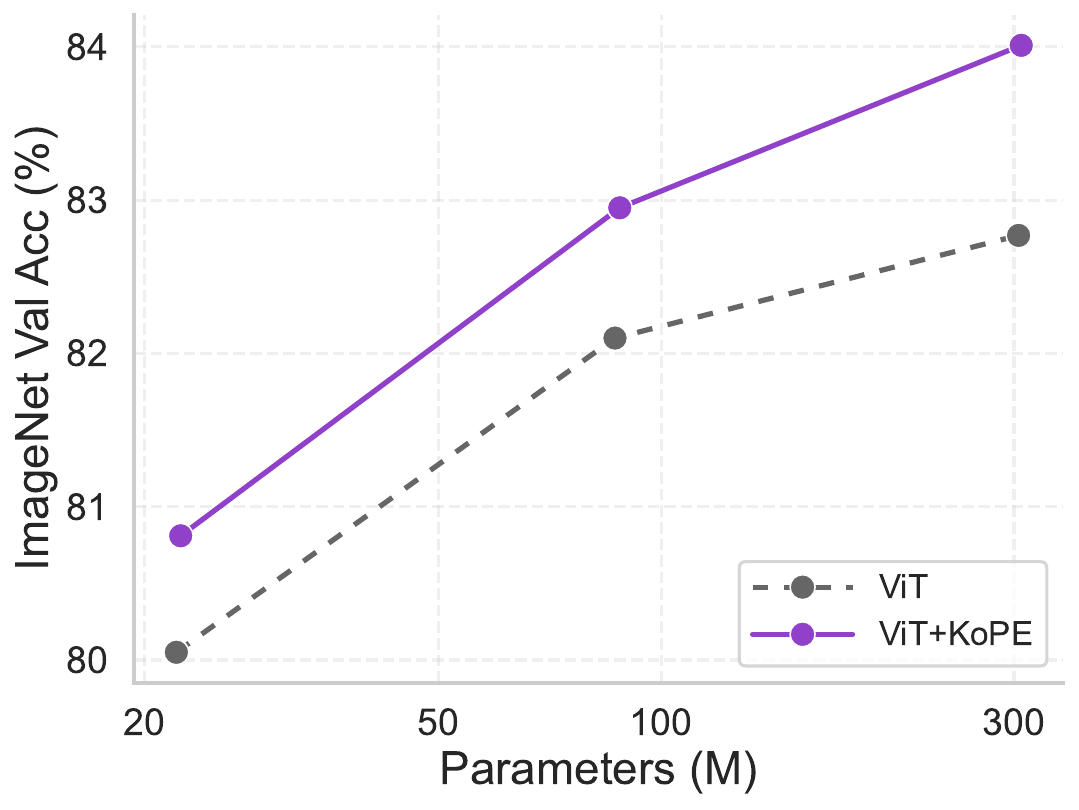}
    \subcaption{Param--Acc (ImageNet val)}
    \label{fig:params-in-val}
  \end{subfigure}
  \hfill
  \begin{subfigure}[t]{0.24\textwidth}
    \centering
    \includegraphics[width=\linewidth]{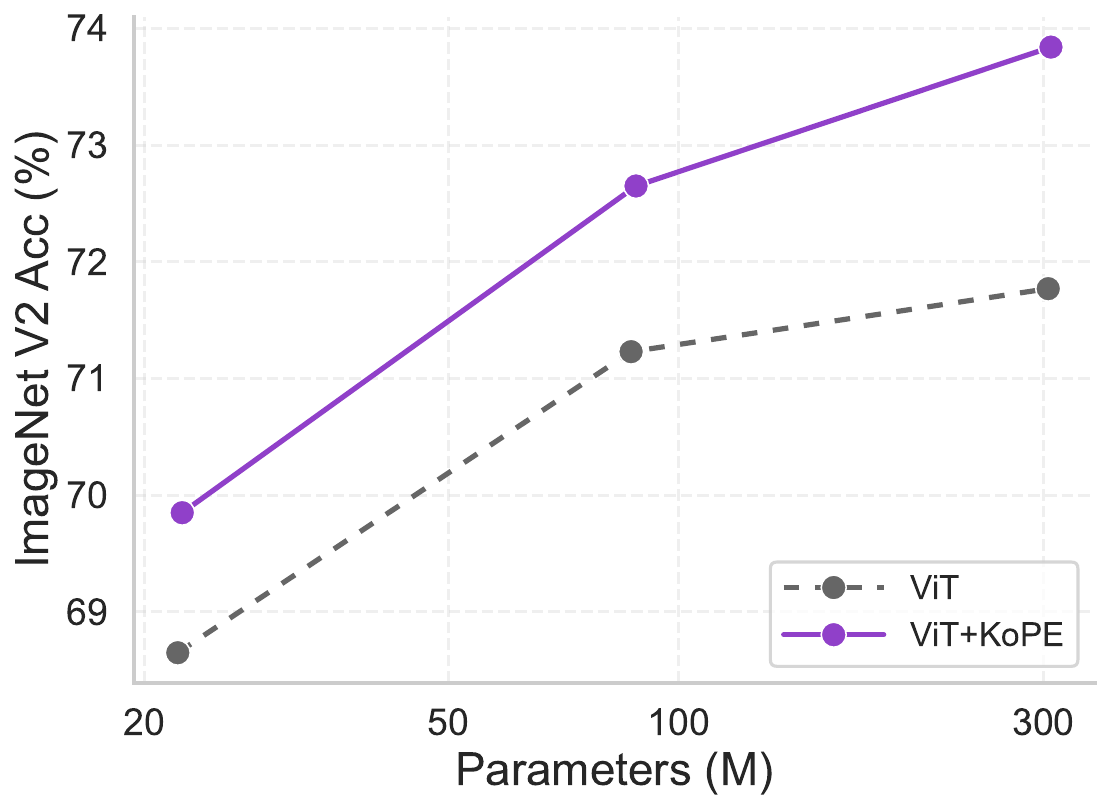}
    \subcaption{Param--Acc (ImageNet V2)}
    \label{fig:params-inv2}
  \end{subfigure}
  \hfill
  \begin{subfigure}[t]{0.24\textwidth}
    \centering
    \includegraphics[width=\linewidth]{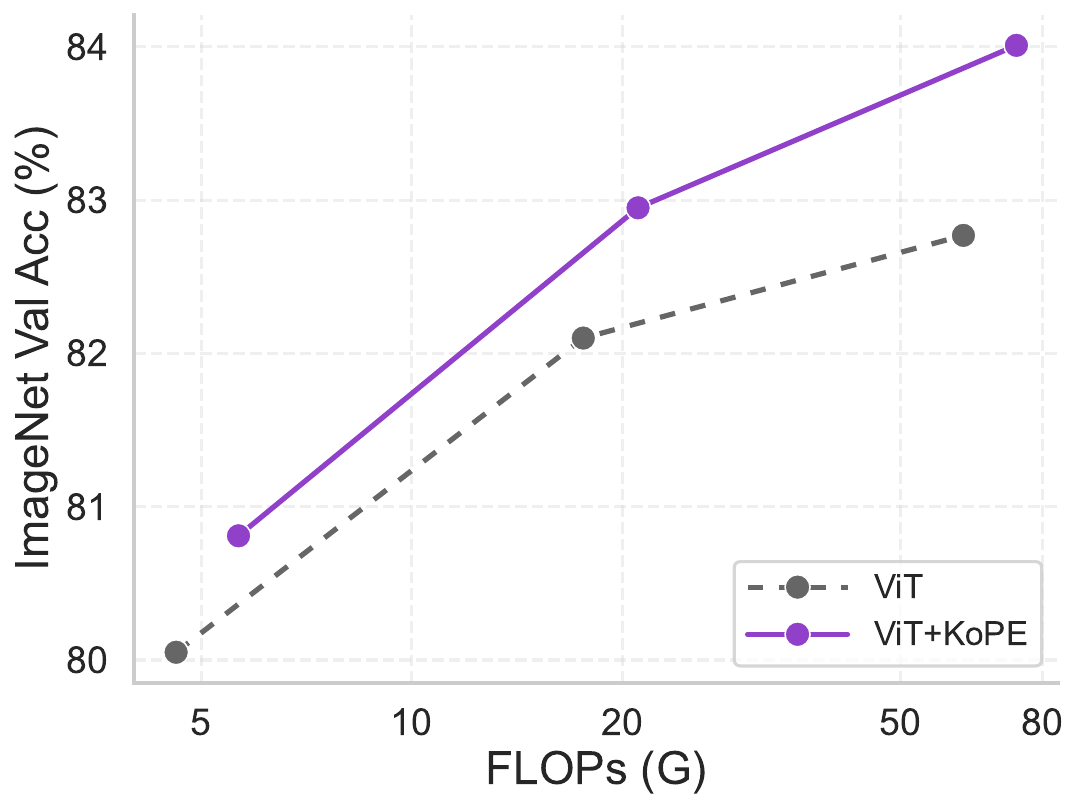}
    \subcaption{FLOPs--Acc (ImageNet val)}
    \label{fig:flops-in-val}
  \end{subfigure}
  \hfill
  \begin{subfigure}[t]{0.24\textwidth}
    \centering
    \includegraphics[width=\linewidth]{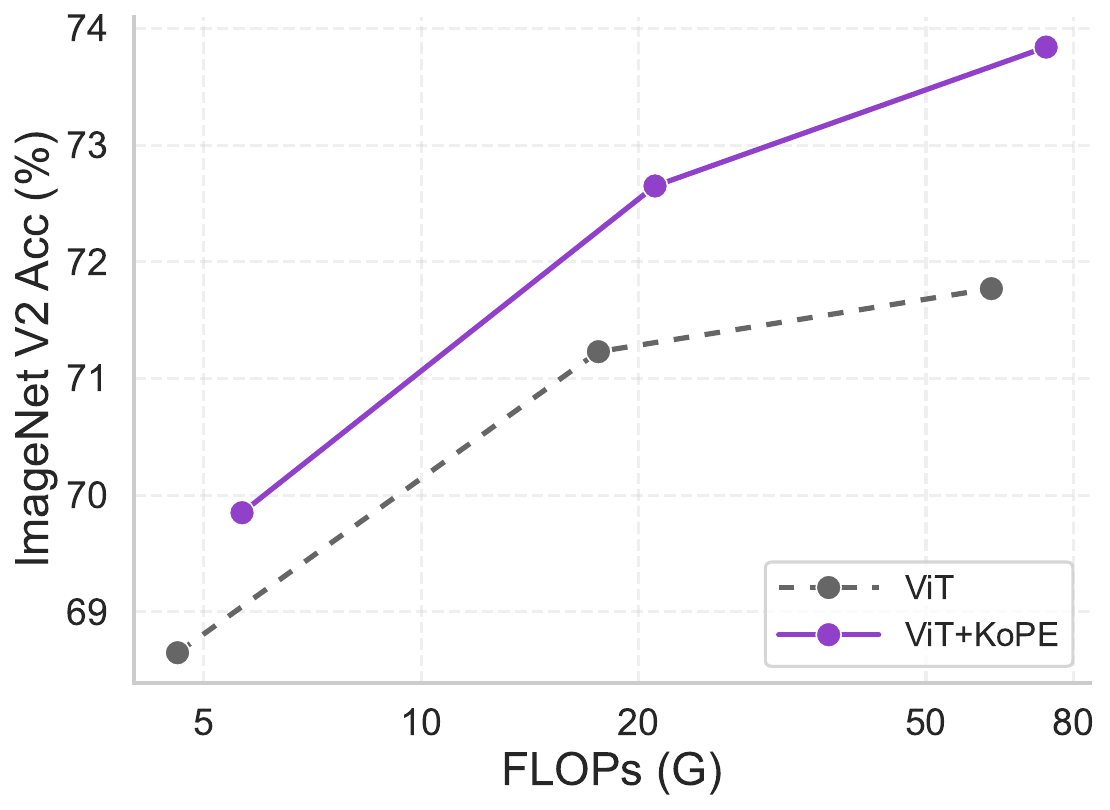}
    \subcaption{FLOPs--Acc (ImageNet V2)}
    \label{fig:flops-inv2}
  \end{subfigure}

  \caption{Summary of learning efficiency results on ImageNet-1K. (a-b) Training dynamics of different models at early and late stages. (c) Accuracy of ViT-B and ViT+KoPE-B trained with different fractions of the training data. (d-e) Parameter--accuracy trade-off curve on ImageNet validation set and ImageNet V2. (f-g) FLOPs--accuracy trade-off curve on ImageNet validation set and ImageNet V2.}
  \label{fig:learning-efficiency}
  %\vspace{-1mm}
\end{figure*}

\subsection{Intuition of KoPE}

KoPE provides a different phase dimension to encode structure information from data through data-adaptive synchronization, facilitating the interactive attention module to formulate structured representations. Intuitively, KoPE aims to synchronize phases from the same entity or concept, and phase differences will encode relations among the entities (e.g., small differences indicate the same entity, or certain angles represent specific relations). Then the attention module is informed by such structure information through phase rotation, learning to adjust attention in a structured way.

This decoupled dimension benefits (1) maintenance of the structure information without re-learning at each layer, (2) disentangle part of relational learning from attention, so that attention can learn to concentrate faster and is alleviated from learning various different functions (e.g., both entity identification and compositional integration). 
With synchronization-enhanced structure learning, the learning efficiency of the model could be improved.

We will illustrate learning efficiency and benefits for structured understanding in Section~\ref{sec:experiments}, and provide theoretical and empirical analysis for speedup of attention learning and structured attention representations in Section~\ref{sec:analysis}.

\subsection{Implementation Details}

\paragraph{Phase Initialization}
In practice, we initialize phases at the first layer with multi-frequency 2d positional embeddings~\citep{heo2024rotary} so that phase rotation is meaningful in the beginning layers. 
In this case, our phase dynamics start from spatial positions and gradually evolve to semantic-relationally synchronized states. 
Under this setting, we can also adopt a mixture of phases in attention calculation. Please refer to Appendix~\ref{supsec:implementation_details} for details. We also verified the importance of a meaningful phase initialization in Appendix~\ref{supsec:more-analysis}.

%\vspace{-2mm}
\paragraph{Parameter and Computation}
We parameterize $h_q$ and $h_k$ by linear projections. Since they are shared across layers, it introduces little additional parameters. $\gamma$ is usually a fixed scalar and we can also parameterize it. The overall parameter overhead is only 1-2\% of original parameters. For the computation of phase update, we can reuse the existing efficient attention calculation methods. It brings about 20\% additional FLOPs for the common setting of ViTs. We can also control similar FLOPs by reducing the network (Appendix~\ref{supsec:match-flops}).

\section{Experiments}\label{sec:experiments}

In this section, we conduct extensive experiments among supervised learning and self-supervised learning (SSL) to illustrate the learning efficiency of KoPE and its superior performance for tasks demanding structured understanding, including visual segmentation, vision-language learning, and few-shot abstract visual reasoning. We first start from the conventional supervised learning on ImageNet-1K~\citep{deng2009imagenet} to demonstrate the learning efficiency.

\subsection{Learning Efficiency}

We train ViT and ViT+KoPE models with different scales (small, base, and large, all with pixel size 16) under the same setting and compare their training dynamics and performance over parameters/FLOPs. We mainly follow the DeiT-III setting~\citep{touvron2022deit} and train models for 300 epochs. As shown in Fig.~\ref{fig:learning-efficiency}(a-b), KoPE largely speeds up the convergence of ViTs under different scales, and the performance gains maintain throughout the training. Considering the final performance, ViT+KoPE achieves better accuracy and show more parameter and computation efficiency as well as larger generalization performance gains on ImageNet V2 (Fig.~\ref{fig:learning-efficiency}(d-g); additional experiments with reduced parameters and matched FLOPs are in Appendix~\ref{supsec:match-flops}). Such gains maintain when we train models for a longer time, e.g., 800 epochs (Appendix~\ref{supsec:longer}), demonstrating the persistent advantages.

We also investigate data efficiency by using different ratio of training data with similar training computing (Fig.~\ref{fig:learning-efficiency}(c)), where ViT+KoPE also shows improved efficiency, achieving similar performance as ViT with 20\% reduction of data.

\begin{figure}
    \centering
    \begin{subfigure}[t]{0.235\textwidth}
      \centering
      \includegraphics[width=\linewidth]{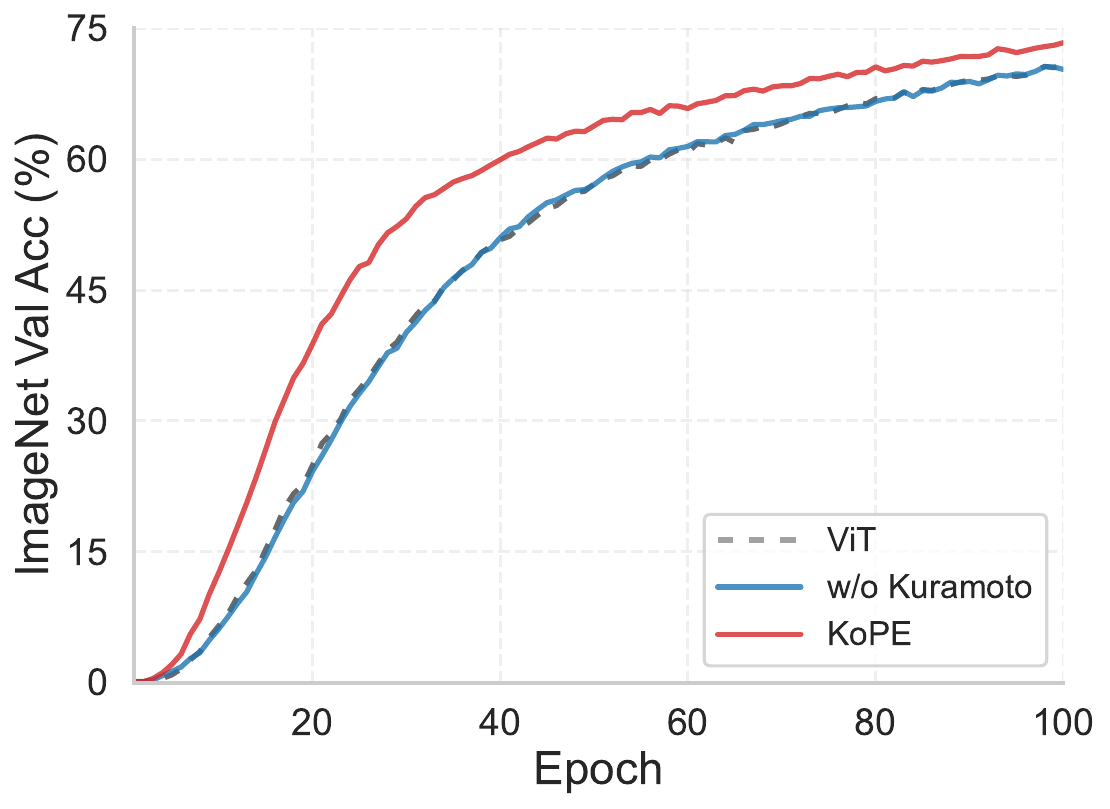}
      \subcaption{Early stage}
    \end{subfigure}
    \hfill
    \begin{subfigure}[t]{0.235\textwidth}
      \centering
      \includegraphics[width=\linewidth]{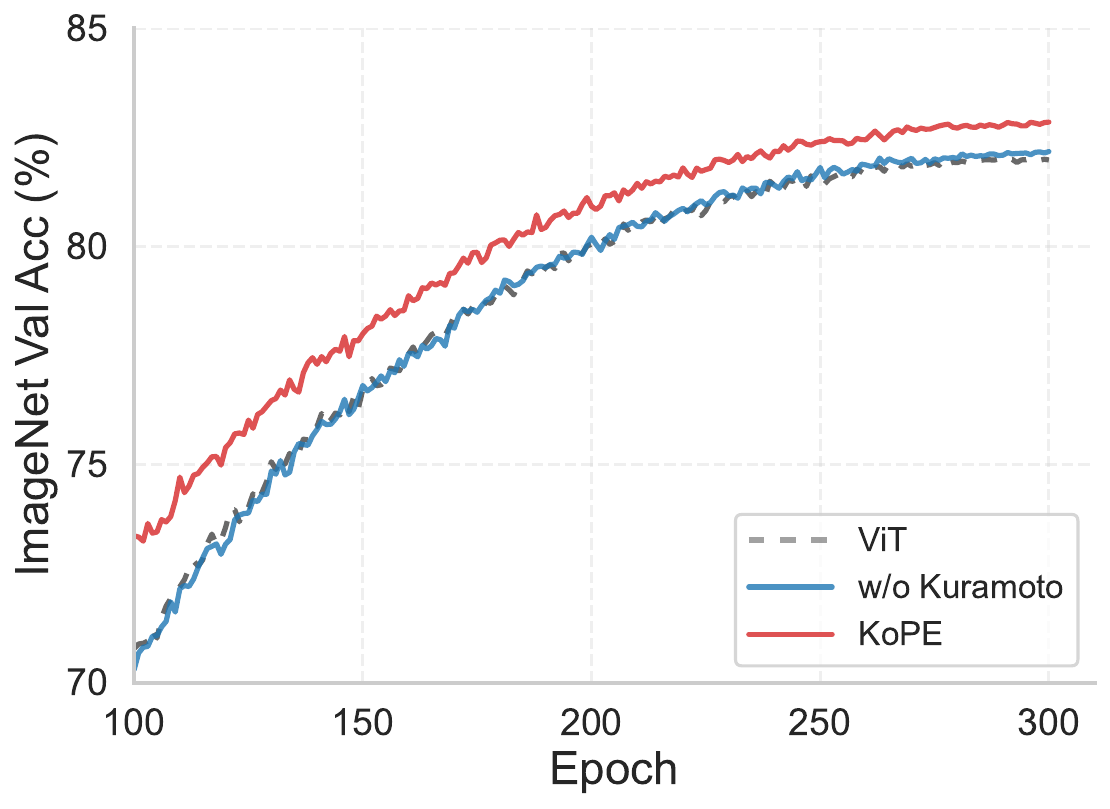}
      \subcaption{Late stage}
    \end{subfigure}
    \caption{Ablation study on learning efficiency. ``w/o Kuramoto'' uses fixed phase states as initialization.}
    \label{fig:ablation}
    %\vspace{-2mm}
\end{figure}

\paragraph{Ablation study}
Fig.~\ref{fig:ablation} verifies that the efficiency comes from Kuramoto dynamics rather than attention rotation, as rotation alone has no improvement over ViT. More analysis experiments can be found in Appendix~\ref{supsec:ablation} and Appendix~\ref{supsec:more-analysis}.

\begin{table}[tb]
  \centering
  \caption{Self-supervised learning results on ImageNet-1K. We train all models by SimDINOv2 for 100 epochs and evaluate linear probing and finetuning accuracy (\%) results.}
  \label{tab:self-supervised}
  \setlength{\tabcolsep}{6.5pt}

  \begin{tabular}{c l c c c c}
    \toprule
    \multirow{2}{*}{\textbf{Scale}} &
    \multirow{2}{*}{\textbf{Model}} &
    \multirow{2}{*}{\textbf{Linear}} &
    \multicolumn{3}{c}{\textbf{Finetuning}} \\
    \cmidrule(lr){4-6}
    & & & \textbf{val} & \textbf{ReaL} & \textbf{V2} \\
    \midrule

    \multirow{2}{*}{small} &
    ViT & 76.9 & 81.2 & 87.1 & 70.7 \\
    & ViT+KoPE & \textbf{77.5} & \textbf{81.7} & \textbf{87.3} & \textbf{71.7} \\
    \midrule

    \multirow{2}{*}{base} &
    ViT & 80.7 & 82.7 & 87.2 & 72.2 \\
    & ViT+KoPE & \textbf{80.9} & \textbf{83.2} & \textbf{87.5} & \textbf{73.5} \\
    \midrule

    \multirow{2}{*}{large} &
    ViT & 82.6 & 83.4 & 87.7 & 73.3 \\
    & ViT+KoPE & \textbf{82.7} & \textbf{84.1} & \textbf{88.3} & \textbf{73.8} \\
    \bottomrule
  \end{tabular}
  %\vspace{-2mm}
\end{table}

\subsection{Self-supervised Learning}

Modern visual models leverage self-supervised learning for scaling~\citep{caron2021emerging,oquab2024dinov,simeoni2025dinov3}, and we show the compatibility of the proposed KoPE with SSL. We consider SimDINOv2~\citep{wusimplifying}, which is a simplified yet better variant of DINOv2~\citep{oquab2024dinov} based on coding rate regularization. We train the models on ImageNet-1K for 100 epochs and evaluate the linear probing and finetuning performance on ImageNet.

As shown in Table~\ref{tab:self-supervised}, KoPE consistently improves the performance of ViT under different scales, verifying the architectural improvement for state-of-the-art learning paradigms. 

\begin{table}[t]
  \caption{Semantic segmentation results on ADE20K (val) with SETR-PUP under single-scale evaluation.}
  \label{tab:ade20k}
  \centering
  %\small
  \begin{tabular}{lcc}
    \toprule
    \textbf{Backbone} & \textbf{mIoU (\%) $\uparrow$} & \textbf{PixAcc (\%) $\uparrow$} \\
    \midrule
    ViT-B & 44.10 & 81.58 \\
    ViT+KoPE-B & \textbf{46.94} & \textbf{82.45} \\
    \midrule
    ViT-L & 47.48 & 82.78 \\
    ViT+KoPE-L & \textbf{48.84} & \textbf{83.04} \\
    \bottomrule
  \end{tabular}
\end{table}

\begin{table}[t]
  \caption{Panoptic segmentation results on COCO (val2017) with Mask2Former under single-scale evaluation.}
  \label{tab:coco_panoptic}
  \centering
  %\small
  \setlength{\tabcolsep}{4.5pt}
  \begin{tabular}{lccccc}
    \toprule
    \textbf{Backbone} &
    \textbf{PQ $\uparrow$} &
    \textbf{PQ$_\textbf{th}$ $\uparrow$} &
    \textbf{PQ$_\textbf{st}$ $\uparrow$} &
    \textbf{SQ $\uparrow$} &
    \textbf{RQ $\uparrow$} \\
    \midrule
    ViT-B      & 54.51 & 60.34 & 45.72 & 83.18 & 64.75 \\
    ViT+KoPE-B & \textbf{55.53} & \textbf{61.67} & \textbf{46.25} & \textbf{83.32} & \textbf{65.82} \\
    \bottomrule
  \end{tabular}
\end{table}

\begin{figure*}[t]
  \centering
  \begin{subfigure}[t]{0.15\textwidth}
    \centering
    \includegraphics[width=\linewidth]{000000000776_img.jpg}
    \subcaption{Original image}
    \label{fig:seg_img1}
  \end{subfigure}
  \hfill
  \begin{subfigure}[t]{0.15\textwidth}
    \centering
    \includegraphics[width=\linewidth]{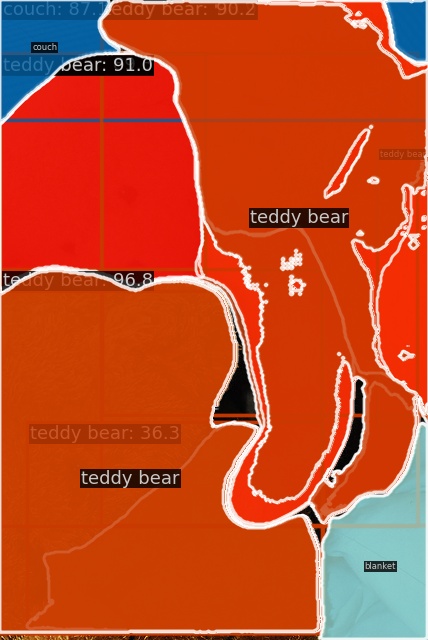}
    \subcaption{ViT-B}
    \label{fig:seg_vit1}
  \end{subfigure}
  \hfill
  \begin{subfigure}[t]{0.15\textwidth}
    \centering
    \includegraphics[width=\linewidth]{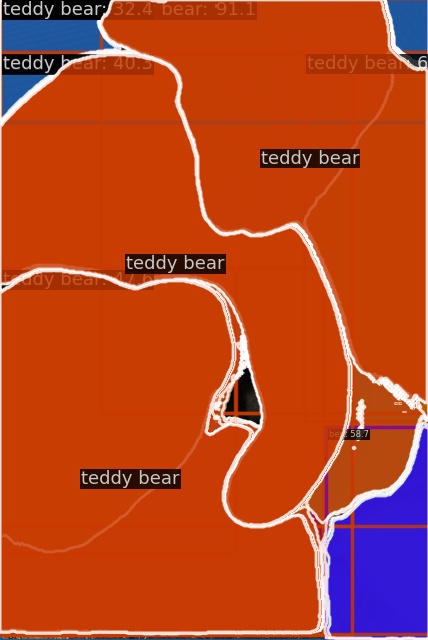}
    \subcaption{ViT+KoPE-B}
    \label{fig:seg_vit_kope1}
  \end{subfigure}
  \hfill
  \begin{subfigure}[t]{0.15\textwidth}
    \centering
    \includegraphics[width=\linewidth]{000000291634_img.jpg}
    \subcaption{Original image}
    \label{fig:seg_img2}
  \end{subfigure}
  \hfill
  \begin{subfigure}[t]{0.15\textwidth}
    \centering
    \includegraphics[width=\linewidth]{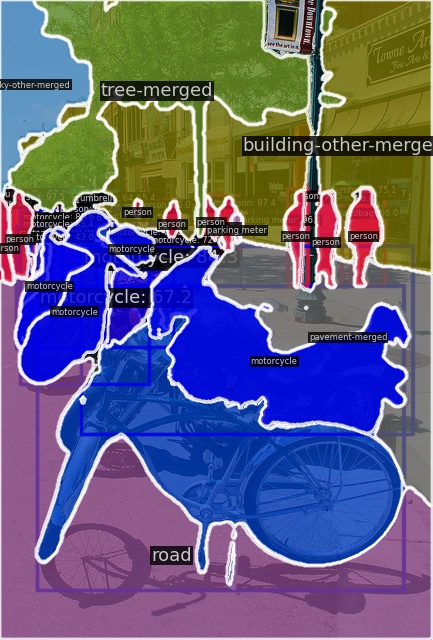}
    \subcaption{ViT-B}
    \label{fig:seg_vit2}
  \end{subfigure}
  \hfill
  \begin{subfigure}[t]{0.15\textwidth}
    \centering
    \includegraphics[width=\linewidth]{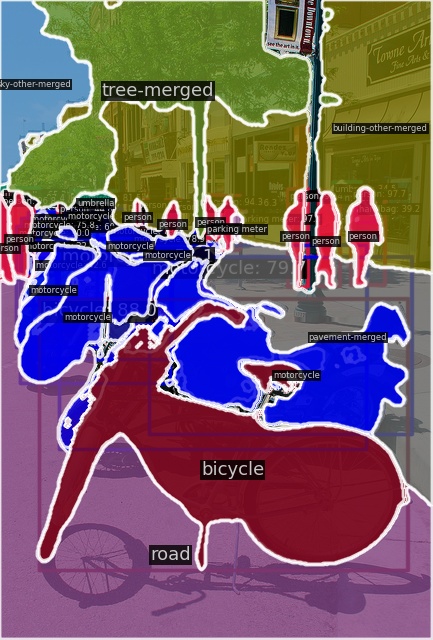}
    \subcaption{ViT+KoPE-B}
    \label{fig:seg_vit_kope2}
  \end{subfigure}

  \caption{Visualization of panoptic segmentation results with Mask2Former under ViT and ViT+KoPE backbones.}
  %\vspace{-1mm}
  \label{fig:visualize_panoptic}
\end{figure*}

\subsection{Semantic and Panoptic Segmentation}

Since phase synchronization provides a potential inductive bias for binding, we expect it to benefit visual segmentation which requires identifying and binding concepts or objects. We evaluate the model on two tasks: (1) semantic segmentation on ADE-20K~\citep{zhou2019semantic}; 
(2) panoptic segmentation on COCO~\citep{kirillov2019panoptic}, which is a more challenging task that jointly unifies instance and semantic segmentation to produce a complete scene parse. For semantic segmentation, we leverage the SETR-PUP~\citep{zheng2021rethinking} that builds light heads on the ViT backbones and report mIoU and pixel accuracy. For panoptic segmentation, we adopt the state-of-the-art Mask2Former~\citep{cheng2022masked} and use ViT or ViT+KoPE as backbones with a simple feature-to-pyramid module on the last layer similar to \citet{li2022exploring}. We report panoptic quality for all, things, and stuff (PQ, PQ$_\text{th}$, PQ$_\text{st}$), segmentation quality (SQ), and recognition quality (RQ). We use backbones pretrained by SimDINOv2 for 100 epochs on ImageNet-1K. More details can be found in Appendix~\ref{supsec:implementation_details}.

As shown in Table~\ref{tab:ade20k} and Table~\ref{tab:coco_panoptic}, KoPE largely improves ViT for the segmentation results, and the advantage of KoPE is amplified on a complex subset (Appendix~\ref{supsec:more-analysis}). The learning efficiency is also improved (see Appendix~\ref{supsec:dynamics-segmentation}). Visualization results are presented in Fig.~\ref{fig:visualize_panoptic}, showing that KoPE encourages more correct binding of parts so that objects can be successfully detected and segmented. This verifies the benefit of phase synchronization for binding in real large-scale settings.

\begin{figure}[t]
  \centering
  \begin{subfigure}[t]{0.235\textwidth}
    \centering
    \includegraphics[width=\linewidth]{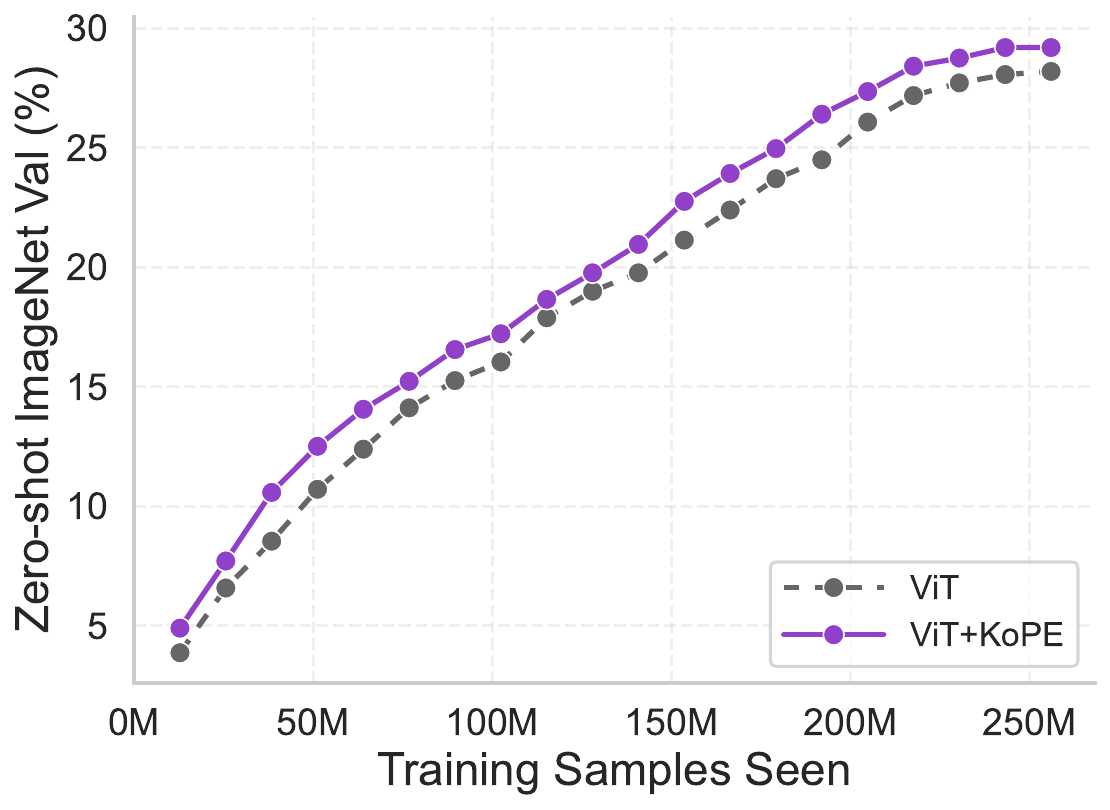}
    \subcaption{ImageNet val}
    \label{fig:clip_imagenet_val}
  \end{subfigure}
  %\hfill
  \begin{subfigure}[t]{0.235\textwidth}
    \centering
    \includegraphics[width=\linewidth]{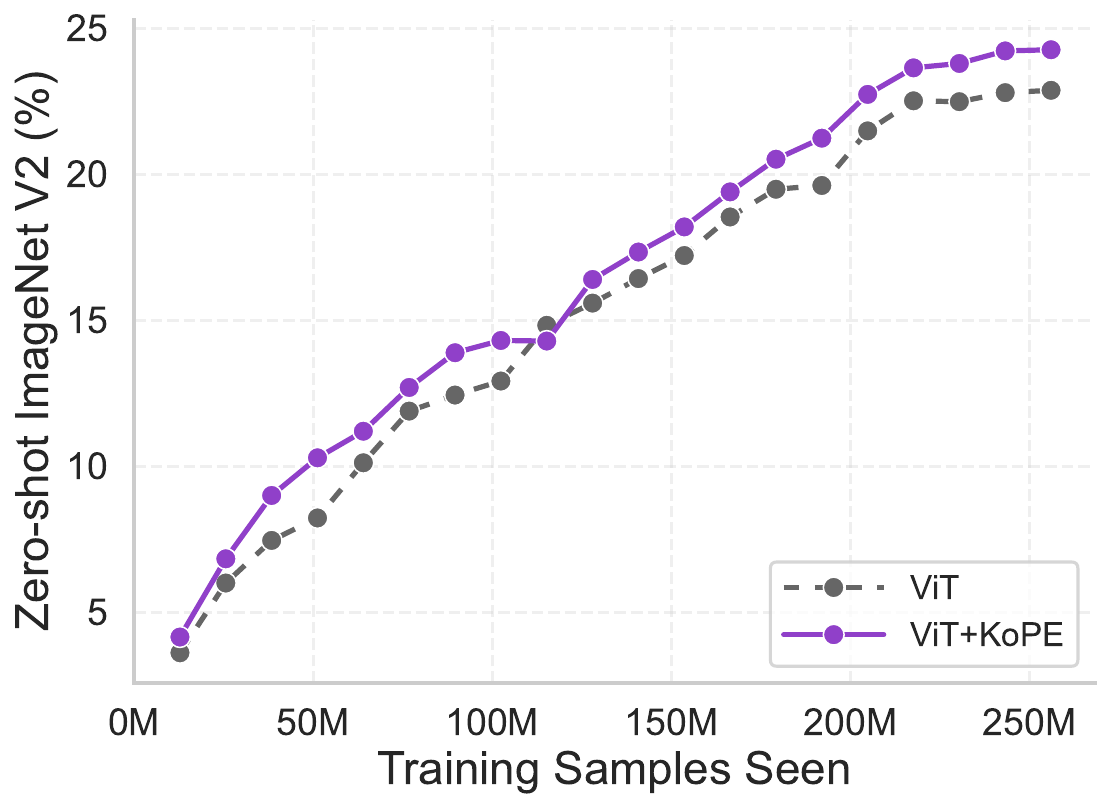}
    \subcaption{ImageNet V2}
    \label{fig:clip_imagenet_v2}
  \end{subfigure}

  \caption{Training dynamics comparison between ViT-B and ViT+KoPE-B in vision-language learning, evaluated by zero-shot classification performance on ImageNet validation set and ImageNet V2.}
  \label{fig:clip}
  %\vspace{-2mm}
\end{figure}

\begin{table}[t]
  \caption{Summary of zero-shot generalization performance under vision-language learning. ``Robust'' is the average of ImageNet V2/A/R/Sketch/O and ObjectNet. ``VTAB'' is the average of visual task adaptation benchmark tasks. ``Retrv.'' is the average Recall@5 among Flickr30k, Flickr8k, and MSCOCO.}
  \label{tab:clip_generalization}
  \centering
  \small
  \setlength{\tabcolsep}{2.5pt}
  \begin{tabular}{lccccc}
    \toprule
    \textbf{Model} &
    \textbf{IN-1K} &
    \textbf{Robust} &
    \textbf{VTAB} &
    \textbf{Retrv. I2T} &
    \textbf{Retrv. T2I} \\
    \midrule
    ViT-B      & 28.21 & 23.44 & 24.50 & 41.62 & 54.59 \\
    ViT+KoPE-B & \textbf{29.22} & \textbf{24.88} & \textbf{28.25} & \textbf{42.45} & \textbf{55.58} \\
    \bottomrule
  \end{tabular}
  %\vspace{-2mm}
\end{table}

\subsection{Vision-language Learning}

Recent advancements in multi-modal models increasingly pose the requirement to align vision models with languages. Since languages are highly abstract and conceptual, we also expect KoPE to facilitate alignment learning with languages. We evaluate ViT and ViT+KoPE under CLIP-style learning~\citep{radford2021learning}, using the medium-scale data from DataComp~\citep{gadre2023datacomp} and the OpenCLIP framework~\citep{ilharco_gabriel_2021_5143773,cherti2023reproducible}. During training, we evaluate the zero-shot classification performance on ImageNet validation set and ImageNet V2; after training, we also systematically evaluate the zero-shot performance on 40 datasets following CLIP benchmark~\citep{cherti_clip_bench}. More details can be found in Appendix~\ref{supsec:implementation_details}.

\textbf{Learning efficiency}\quad As shown in Fig.~\ref{fig:clip}, KoPE improves the learning efficiency and final performance of vision-language models, achieving similar performance with 15-20\% less seen training samples. 

\textbf{Generalization}\quad Meanwhile, ViT+KoPE has better zero-shot generalization performance (Table~\ref{tab:clip_generalization} and Appendix~\ref{supsec:clip}), with remarkable robust generalization performance (e.g., +2.69 pp on ObjectNet, indicating that KoPE encourages more object-centric recognition) and adaptation performance (+3.75 pp in average). These results further suggests the (symbolic) structure learning ability of KoPE.

\begin{table}[t]
    \centering
    \small
    \caption{Results on ARC-AGI tasks. We pretrain the model and perform test-time training following~\citet{hu2025arc}, and report mean$\pm$std (max) over five runs of test-time training. * denotes continual pretraining on ARC-AGI-2 training tasks (VARC default on ARC-AGI-1). Comparison results taken from~\citet{hu2025arc}.}
    \label{tab:arc_results}
    
    \definecolor{groupgray}{gray}{0.93}
    \definecolor{rowtextgray}{gray}{0.60}
    
    \setlength{\tabcolsep}{1pt}
    
    \begin{tabular}{l c c c}
    \toprule
    \textbf{system} & \textbf{\#params} & \textbf{ARC-1} & \textbf{ARC-2} \\
    \midrule
    
    \rowcolor{groupgray}
    \multicolumn{4}{l}{\itshape large language models (LLMs)} \\
    {\color{rowtextgray}Deepseek R1}      & {\color{rowtextgray}671B} & {\color{rowtextgray}15.8} & {\color{rowtextgray}1.3} \\
    {\color{rowtextgray}Claude 3.7 8k}    & {\color{rowtextgray}N/A}  & {\color{rowtextgray}21.2} & {\color{rowtextgray}0.9} \\
    {\color{rowtextgray}o3-mini-high}     & {\color{rowtextgray}N/A}  & {\color{rowtextgray}34.5} & {\color{rowtextgray}3.0} \\
    {\color{rowtextgray}GPT-5}            & {\color{rowtextgray}N/A}  & {\color{rowtextgray}44.0} & {\color{rowtextgray}1.9} \\
    {\color{rowtextgray}Grok-4-thinking}  & {\color{rowtextgray}1.7T} & {\color{rowtextgray}66.7} & {\color{rowtextgray}16.0} \\
    {\color{rowtextgray}Bespoke (Grok-4)} & {\color{rowtextgray}1.7T} & {\color{rowtextgray}\textbf{79.6}}              & {\color{rowtextgray}\textbf{29.4}} \\
    \addlinespace[2pt]
    
    \rowcolor{groupgray}
    \multicolumn{4}{l}{\itshape recurrent models} \\
    HRM & 27M & 40.3 & 5.0 \\
    TRM & 7M  & 44.6 & 7.8 \\
    \addlinespace[2pt]
    
    \rowcolor{groupgray}
    \multicolumn{4}{l}{\itshape vision models (VARC)} \\
    ViT~\citep{hu2025arc} & 18M & 54.5$\pm$0.7 & 8.3$\pm$0.4 \\
    ViT~\citep{hu2025arc} & 66M & 53.0 & - \\
    \midrule
    Enc-Dec ViT & 36M & 54.7$\pm$0.6 (55.3) & 7.3$\pm$0.9 (8.3) \\
    Enc-Dec ViT+KoPE & 37M & \textbf{56.8$\pm$0.7 (57.5)} & \textbf{9.8$\pm$0.6 (10.8)} \\
    Enc-Dec ViT+KoPE* & 37M & / & \textbf{10.2$\pm$0.4 (10.8)} \\
    %\addlinespace[2pt]
    
    %\rowcolor{groupgray}
    %\multicolumn{4}{l}{\itshape human results} \\
    %avg.\ human  & -- & 60.2 & -- \\
    %best human   & -- & 98.0 & 100.0 \\
    \bottomrule
    \end{tabular}
    %\vspace{-4mm}
\end{table}

\subsection{Abstract Visual Reasoning}

Finally, we investigate challenging few-shot abstract visual reasoning tasks, ARC-AGI-1~\citep{chollet2019measure} and ARC-AGI-2~\citep{chollet2025arc}, which are easy for humans while hard for state-of-the-art AI models. These tasks require concepts of objectness and compositionality, and we expect that KoPE can improve current models. We follow the VARC method~\citep{hu2025arc} and compare ViT+KoPE versus ViT in an adapted encoder-decoder framework. Please refer to Appendix~\ref{supsec:experiment_details} for details.

As shown in Table~\ref{tab:arc_results}, the encoder-decoder model with ViT+KoPE achieves state-of-the-art performance over single models trained from scratch~\citep{wang2025hierarchical,jolicoeur2025less,hu2025arc}, validating the benefit of KoPE for visual reasoning with structured understanding.

\section{Analysis of KoPE}\label{sec:analysis}

In this section, we further provide analysis to present intuitions of why KoPE improves the learning efficiency of ViTs. Recent theoretical results have shown the important of attention sparsity (concentration) during training evolution~\citep{li2023theoretical} and for emergence~\citep{zucchet2025emergence}, and training/sample complexity of shallow ViTs can be derived~\citep{li2023theoretical}. We first show that in a simplified theoretical setting, KoPE can accelerate attention learning (i.e., the sparse attention to focus on relevant parts) and potentially reduce training and sample complexity, and then provide empirical verification.

\begin{figure}[t]
  \centering
  % ---------------- Row 1: 2 plots ----------------
  \begin{subfigure}[t]{0.235\textwidth}
    \centering
    \includegraphics[width=\linewidth]{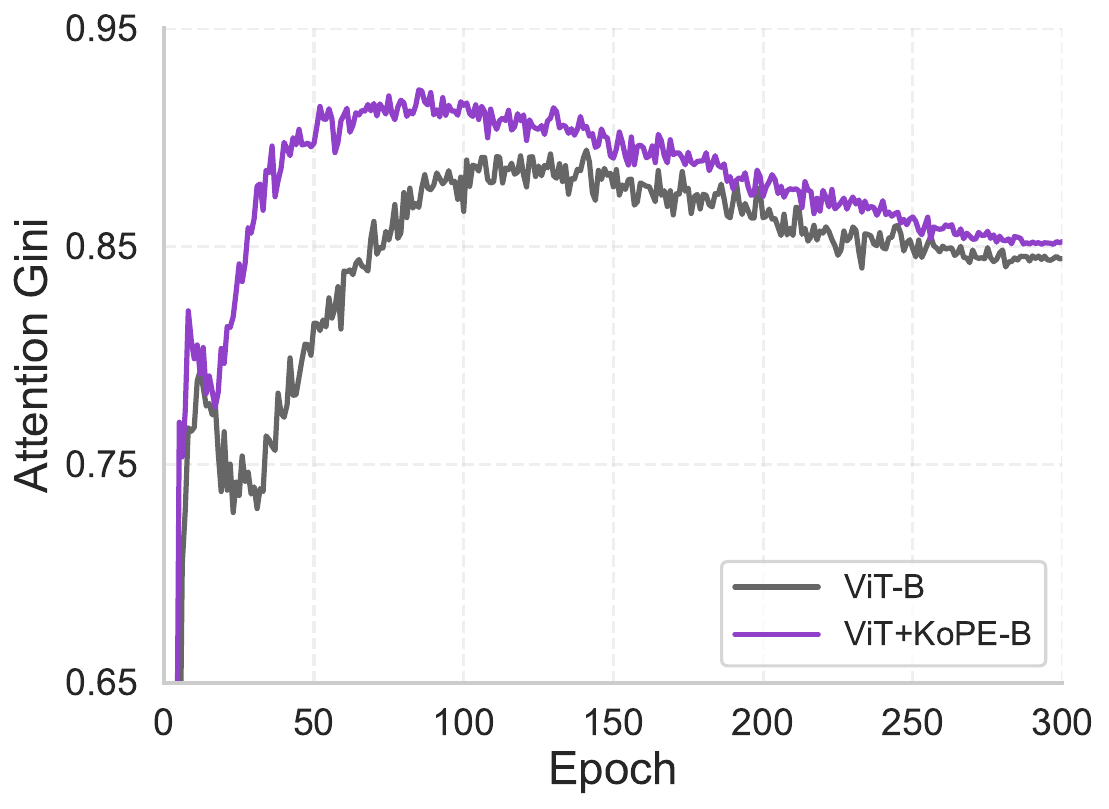}
    \subcaption{Gini metric of attention}
    \label{fig:attn_gini}
  \end{subfigure}
  \hfill
  \begin{subfigure}[t]{0.235\textwidth}
    \centering
    \includegraphics[width=\linewidth]{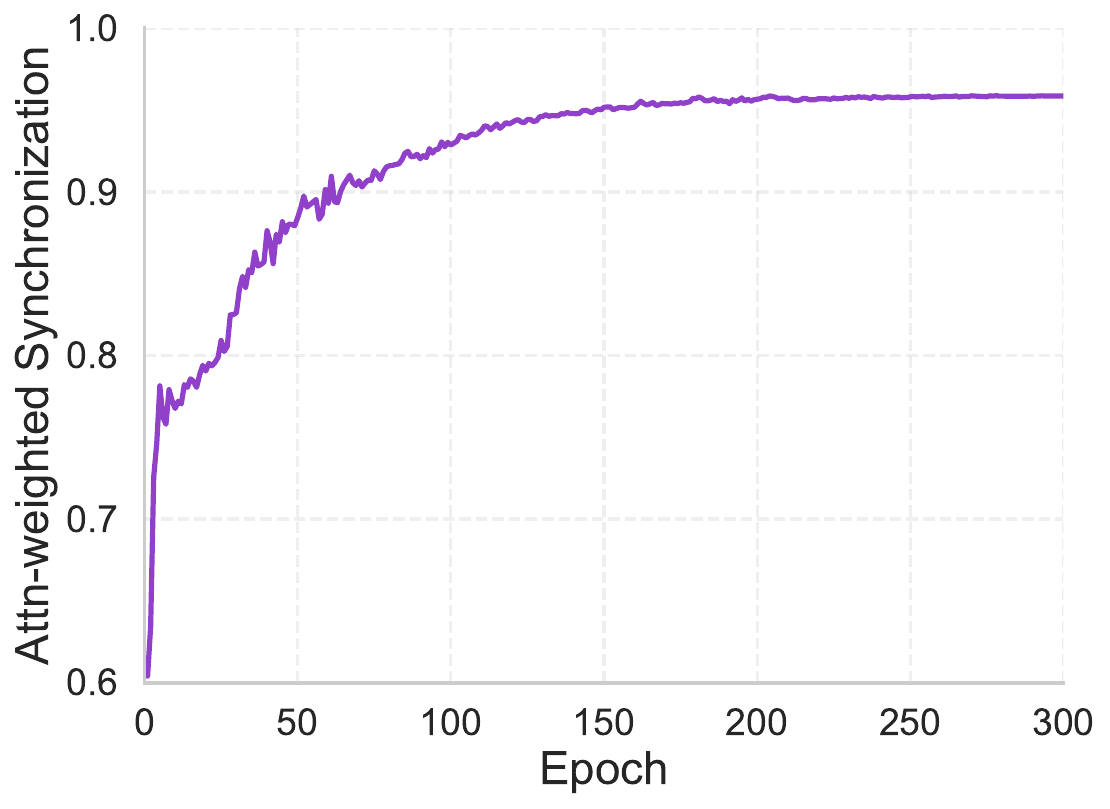}
    \subcaption{Weighted synchronization}
    \label{fig:phase_sync}
  \end{subfigure}

  \caption{Empirical verification of attention concentration and weighted phase synchronization throughout ImageNet supervised learning. (a) Evolution of average Gini metric over all tokens for all heads of the attention of CLS token in the last layer during training. (b) Evolution of phase synchronization weighted by the attention of CLS token in the last layer during training.}
  \label{fig:attention_concentration}
  %\vspace{-2mm}
\end{figure}

\begin{figure}[t]
  \centering
  \includegraphics[width=0.95\linewidth]{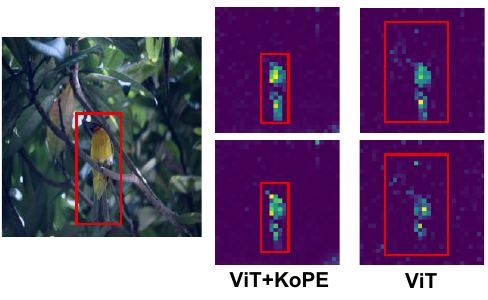}
  \caption{Visualization of attention maps for models (base) trained on ImageNet-1K under supervised learning. Attention maps are from two best attention heads of CLS token in the last layer.}
  \label{fig:visualize_attn}
  %\vspace{-2mm}
\end{figure}

\subsection{Attention Concentration and Training Efficiency}

We first consider a simplified theoretical setting for analysis. As in \citet{li2023theoretical}, we consider binary classification over token sequences with both discriminative and non-discriminative parts, for a shallow ViT with one self-attention layer followed by a two-layer perception. We append a CLS token in the model. For KoPE, we consider the setting that phases have converged to clustered synchronization, analogous to the final layer in real ViTs. The cluster tightness, separation, and consistency will be characterized by $1-\varepsilon_{\phi}/2$, $\Delta\phi_{\text{min}}$, and $1-\xi_\phi$, and we have $\rho_\phi=\cos(\Delta\phi_{\text{min}})<\gamma_\phi=\cos(\varepsilon_\phi)$. We also simplify the rotation in attention as a function to enhance within-cluster interaction and suppress cross-cluster parts. This is only a simplified setting for intuition and is a conservative explanation for KoPE, as it may be more complex and powerful for feature extraction from complex data in real scenarios.
    
We show that under this simplified setting, the condition for attention concentration on relevant token ($\sum_{j\in S_*}a_{0j}$) is relaxed by KoPE, so it is likely accelerated.
\begin{lemma}[Relaxed condition for attention concentration, informal]
    Let $\Delta=\min_{j\in S_*}\langle q, k_j\rangle-\max_{r\in S_{\neg *}}\langle q, k_r\rangle$ denote the content-only logit gap for relevant tokens from the CLS token. The condition to reach $1-\epsilon$ attention concentration on relevant tokens is relaxed from $\Delta\geq \log\frac{|S_{\neg *}|}{|S_*|}+\log\frac{1}{\epsilon}$ to $\Delta\geq \log\frac{|S_{\neg *}|}{|S_*|}+\log\frac{1}{\epsilon}-f(\gamma_\phi,\rho_\phi, \xi_\phi)$, where $f(\gamma_\phi,\rho_\phi, \xi_\phi)>0$ is a function depending on $\gamma_\phi,\rho_\phi, \xi_\phi$. Then under similar training dynamics that increases $\Delta$, attention concentration will be achieved earlier.
\end{lemma}

%\vspace{-1mm}
Detailed descriptions can be found in Appendix~\ref{supsec:theory_details}.
With more concentrated attention on relevant tokens and suppression of irrelevant information, the training and sample complexity are potentially reduced, given the close relationship between attention concentration and training/sample complexity~\citep{li2023theoretical}.

%\vspace{-1mm}
\subsection{Empirical Verification}

We provide empirical verification of accelerated attention learning by presenting the dynamics of the average Gini metric over all tokens for all heads of the attention from the last layer's CLS token, which measures the concentration of attention. As shown in Fig.~\ref{fig:attention_concentration}(a), KoPE largely accelerates the process of attention concentration, consistent with the theoretical analysis. As the average Gini metric is over all tokens rather than label-relevant tokens, it is only a surrogate for the expected concentration and there exists a two-stage evolution where the first stage increases Gini metric while the second stage decreases it, potentially learning more accurate relevant tokens. KoPE has a larger concentration throughout the training. Fig.~\ref{fig:attention_concentration}(b) also verifies the phase synchronization of attended tokens in the last layer during training (synchronization weighted by the attention from CLS token), indicating both synchronization of relevant parts and the increased attention on them (more analysis results can be found in Appendix~\ref{supsec:more-analysis}). We further visualize the (phase-gated) attention maps from the last layers' CLS token in Fig.~\ref{fig:visualize_attn}, demonstrating that KoPE enables better attention to objects relevant to the labels, suggesting a more structured attention representation on relevant tokens. 
More visualization results in Appendix~\ref{supsec:visualization} further verify the structured representations and the complementary advantages of KoPE over learning paradigms.

%\vspace{-2mm}
\section{Related Work}
%\vspace{-1mm}

A line of works attempt to introduce complex value with phase information into neural networks. Complex-valued neurons~\citep{reichert2013neuronal,lowe2022complex,lee2022complex,lowe2023rotating} introduce complex representations for neural networks and have been applied to object discovery or naturally complex-valued data and signals. Neural synchrony are studied in this context~\citep{reichert2013neuronal,lowe2022complex,stanic2023contrastive,lowe2023rotating} or other spatiotemporal structure such as spiking time~\citep{zheng2022dance,zheng2023gust}. These works mainly focus on binding-specific problems, e.g., object discovery, not targeting general vision tasks. Some recent works introduce Kuramoto dynamics into deep learning~\citep{miyatoartificial,songkuramoto}, formulating phase-only neurons for object discovery, adversarial robustness, and sudoku reasoning~\citep{miyatoartificial}, or injecting structural inductive bias for the diffusion process~\citep{songkuramoto}. Kuramoto model is also introduced in graph learning~\citep{nguyen2024coupled,ding2025let} for the over-smoothing problem. Nevertheless, most works are usually restricted to specific tasks rather than advancing scalable modern neural networks for large-scale vision tasks. In contrast, the proposed KoPE instantiates coupled rate--phase dynamics within standard Vision Transformers, enabling synchronization as an structural inductive bias under realistic large-scale training.

Complex formulation has also been studied in the literature of positional embedding, with the representative method of rotary positional embedding~\citep{su2024roformer} and subsequent extensions to 2D positions~\citep{heo2024rotary}, longer contexts~\citep{pengyarn}, or 3D geometry~\citep{miyato2024gta}. Nevertheless, they focus on static phases with only positional/geometric information, while we investigate evolving phases driven by Kuramoto dynamics to learn structure from data through synchronization.

Besides neural synchrony, the binding problem has been studied in the object-centric literature, which is regarded as a key obstacle for compositional and systematic generalization~\cite{greff2020binding}. Slot-based models~\citep{burgess2019monet,greff2019multi,locatello2020object} are popular methods for object-centric learning. However, similar to previous synchrony-based models, they struggle on scalability to natural images and require combination with pre-trained vision models~\citep{seitzer2023bridging}. The proposed KoPE, on the other hand, shows that synchronization can be leveraged as a scalable structural inductive bias for large-scale vision tasks with implicit synchronization-based binding instead of explicit slots. This re-positions neural synchrony as an important and scalable mechanism for solving this problem.

A broad literature studies learning efficiency in vision through architectural/connectivity design~\citep{liu2021swin}, distillation or training recipe~\citep{touvron2021training,touvron2022deit}, data pruning~\citep{paul2021deep}, optimization~\citep{foretsharpness,xie2024adan}, etc. The proposed KoPE turns to a different perspective on neural representation and information processing through phases, providing a structural inductive bias that is compatible with most of these works. 

%\vspace{-1mm}
\section{Conclusion}

In this work, we propose Kuramoto oscillatory Phase Encoding, an evolving phase state incorporated into the mainstream Vision Transformer model. We show that neuro-inspired synchronization can advance scalable neural networks for training, parameter, and data efficiency, meanwhile benefiting tasks requiring structured understanding such as image segmentation, vision-language alignment, and few-shot abstract visual reasoning. Extensive experiments and theoretical analysis verify the efficiency and superior performance of the proposed method. Our work shows that synchronization can serve as a scalable, neuro-inspired mechanism for advancing state-of-the-art neural networks, shedding light on new routes for bridging brain-inspired principles with scalable vision learning.

%\section*{Acknowledgements}

\section*{Impact Statement}

%This paper presents work whose goal is to advance the field of Machine Learning. There are many potential societal consequences of our work, none which we feel must be specifically highlighted here.

This work proposes KoPE as a scalable neuro-inspired mechanism for improving learning efficiency and structured visual understanding. Such mechanisms may help reduce the computation or model capacity needed to reach a target performance level, which could lower the resource barrier for strong vision models. At the same time, KoPE is designed for structured representation learning from unstructured data, which may not be assumed to transfer uniformly to all tasks, domains, or applications. Additionally, broader risks such as data bias, shift, or harmful downstream use are not resolved by this mechanism itself and must still be evaluated separately.

\bibliography{reference}
\bibliographystyle{icml2026}

%%%%%%%%%%%%%%%%%%%%%%%%%%%%%%%%%%%%%%%%%%%%%%%%%%%%%%%%%%%%%%%%%%%%%%%%%%%%%%%
%%%%%%%%%%%%%%%%%%%%%%%%%%%%%%%%%%%%%%%%%%%%%%%%%%%%%%%%%%%%%%%%%%%%%%%%%%%%%%%
% APPENDIX
%%%%%%%%%%%%%%%%%%%%%%%%%%%%%%%%%%%%%%%%%%%%%%%%%%%%%%%%%%%%%%%%%%%%%%%%%%%%%%%
%%%%%%%%%%%%%%%%%%%%%%%%%%%%%%%%%%%%%%%%%%%%%%%%%%%%%%%%%%%%%%%%%%%%%%%%%%%%%%%
\newpage
\appendix
\onecolumn
\section{Implementation Details}\label{supsec:implementation_details}

\subsection{Multi-head attention and coupling}

In practice, the attention module will be divided into multiple heads, i.e., the input $\mathbf{z}\in\mathbb{R}^d$ is treated as $\mathbf{z}\in\mathbb{R}^{n_h\times d_h}$, and attention is carried out for each head with dimension $d_h$. KoPE also maintains separate phases for different heads, which are coupled separately and act on the corresponding attention heads. The coupling connections are calculated for each head, based on per-head representations with dimension $d_h$ from $h_q(\mathbf{z})$ and $h_k(\mathbf{z})$. This is similar to the calculation of multi-head attention, while we do not have value/output projection in the coupling.

\subsection{Phase initialization and calculation}

In practice, we initialize phases at the first layer with multi-frequency 2d positional embeddings so that phase rotation is meaningful in the beginning layers. Specifically, it extends rotary position embedding to 2d grids, and the phases are based on the spatial position $(x, y)$. For each head with dimension $d_h$ (should be dividable by 4), half of the dimensions are used for the $x$-axis while the remaining for the $y$-axis. For each axis, the phase representations at index $i$ are calculated by $\cos(i\psi_j)$ and $\sin(i\psi_j)$, where $\psi_j=\text{base}^{-j/(d_h/4)}$ is the frequency that is different for paired dimension index $j$. For the CLS token, it is initialized with $\phi=0$, i.e., $\cos(\phi)=1$ and $\sin(\phi)=0$. 
In this case, our phase dynamics start from spatial positions and gradually evolve to semantic-relationally synchronized states. We can also learn to determine initial phases from token representations, while we use positional initialization in this paper.

\textbf{Mixture of phases}\quad Then for each head, the coupling process uses a same coupling matrix for phases initialized with different spatial frequencies. This may result in different synchronization structures because they are dependent on the initialization condition as well. 
Under this setting, we can also adopt a mixture of phases in attention calculation to better leverage information from different dimensions. Specifically, let $\mathbf{r}_k\in\mathbf{R}^{(d_h/2)\times 2}$ denote the phase representation of the $k$-th head, where $\mathbf{r}_{k,j}\in\mathbb{R}^{2}$ corresponds the phases that are initialized from the $j$-th spatial frequency. We parameterize a mixing matrix $\mathbf{M}_k\in\mathbb{R}^{(d_h/2)\times(d_h/2)}$ (initialized around the identity matrix) for each head to learn to mix different phases for attention rotation. That is, we use $\Pi^{(2)}(\mathbf{M}_k\mathbf{r}_k)$ for phase rotation. This introduces minimal parameter costs, and analysis in Appendix~\ref{supsec:ablation} shows that this leads to slightly better final performance.

The only hyperparameter in KoPE is the step size $\gamma$. For supervised learning, we fix $\gamma=0.05$. We can also make it learnable, i.e., parameterize it with a softplus function to ensure positivity. In other experiments, we basically use the learnable $\gamma$ initialized at $0.05$.

\section{Experiment Details}\label{supsec:experiment_details}

\subsection{Supervised Learning on ImageNet-1K}

For supervised learning, we mainly follow the DeiT-III setting~\citep{touvron2022deit} to adopt strong data augmentation. We set the basic hyper-parameters, such as learning rate, drop path rate (for different model scales), etc., as the recommended value. For computational efficiency, we mainly train models for 300 epochs, while Appendix~\ref{supsec:longer} also shows results with longer training time (800 epochs). We use exponential moving average (EMA) for models and report the final performance of the EMA models.

For the data efficiency experiments, we use a subset of training data (with different ratios) while keeping the whole iteration similar. For example, with 40\% ratio of training data, we train models for $300/(40\%)=750$ epochs to keep similar training steps. This is to study the pure influence of the data without the interference of the training steps.

The common ViTs adopt the absolute position embeddings with learnable parameters added to the patch embeddings. We keep this settings for all models. For the ablation study, we only use the initial phases for rotation without Kuramoto dynamics, which is similar to rotary position embedding (also kept absolute position embeddings, without it the performance is lower by ~2\%), and experiments show that it hardly has difference from vanilla ViTs, verifying the necessity of the synchronization prior from Kuramoto dynamics.

\subsection{Self-supervised Learning}

We leverage SimDINOv2~\citep{wusimplifying} to train all models for 100 epochs. The hyperparameters are set as values in their paper, while we set drop path rate as 0.1 for small and base models (0.2 for large models). For finetuning, we train models for 100 epochs for small and base models while 50 epochs for large models. 
For ViT+KoPE-L, we additionally add a regularization term for phases in the self-supervised objective. Specifically, similar to the ibot loss, we encourage the phases of each token at the last layer from the student network to align with the phases from the teacher network. This both regularizes the phases of tokens from the student network's input and encourages predicting phases from the masked patches.

\subsection{Image Segmentation}

For semantic segmentation on ADE-20K, we leverage the SETR-PUP method~\citep{zheng2021rethinking} that builds light heads on the ViT backbone. This enables better evaluation of the advantages of ViT-style backbones. For panoptic segmentation on COCO, we leverage the Mask2Former method~\citep{cheng2022masked} with ViT or ViT+KoPE as backbones. As Mask2Former requires pyramid inputs, we leverage a simple feature-to-pyramid module on the last layer of ViT or ViT+KoPE, similar to \citet{li2022exploring}. It builds feature maps with resolution scales 4, 2, 1, and 0.5 for the network output through transposed convolutions or max pooling, which is then processed by Mask2Former segmentation heads. All experiments are carried out under the mmsegmentation or mmdetection framework, with segmentation head and training settings as the default configs for SETR-PUP and Mask2Former.

\subsection{Vision-language Learning}

We leverage the OpenCLIP framework~\citep{ilharco_gabriel_2021_5143773,cherti2023reproducible} to train ViT and ViT+KoPE under CLIP-style learning. We utilize the unfiltered medium-scale data from DataComp~\citep{gadre2023datacomp}, which originally contains ~128M data (around 4.5 TB) while we only successfully downloaded ~70\% of the data from the Internet (around 3.2 TB). We use the same data and training settings for both models, and we consider ViT-B-16 and ViT+KoPE-B-16 models. The training setting mainly follows the corresponding medium recipe of DataComp~\citep{gadre2023datacomp} while we double the training samples seen (from 128M to 256M).

During training, we evaluate the zero-shot classification performance on ImageNet validation set and ImageNet V2 following the OpenCLIP realization. After training, we also systematically evaluate the zero-shot performance on 40 datasets following CLIP benchmark~\citep{cherti_clip_bench}. The detailed results are reported in Appendix~\ref{supsec:clip}.

\subsection{Abstract Visual Reasoning}

For few-shot abstract visual reasoning on ARC-AGI tasks, we follow the VARC method~\citep{hu2025arc} that views it a vision problem, and leverage the same procedure to patchify images and perform data augmentation. We follow the training pipeline to first pretrain the model on training tasks from ARC-AGI-1 and re-arc for 100 epochs, and then perform test-time training separately for each task in the validation set of ARC-1/ARC-2. Test-time training is performed for multiple times and we report the mean, standard deviation, and max results.

We adapt the pure ViT in \citet{hu2025arc} into an encoder-decoder architecture, where the encoder focuses on understanding the inputs while the decoder focuses on generating the output answer. We only adopt KoPE for the encoder since KoPE facilitates structured understanding. The encoder is composed of self-attention and MLP layers, while the decoder is composed of self-attention, cross-attention, and MLP layers. \citet{hu2025arc} leverage 2d rotary positional embeddings in ViT, and we replace it by our KoPE in the encoder while keeping it in the decoder. There are task tokens $x_{t_i}$ appended to the input tokens in the original setting, and we also keep this setting and leverage task tokens to pass information from encoder to decoder. Specifically, we append the task tokens to inputs for the encoder, which is updated till the output $\hat{x}_{t_i}$, and these tokens for the input to the decoder is set as $x_{t_i}+\hat{x}_{t_i}$. The decoder then uses cross-attention to extract information from the output of the encoder to generate answers. Different from ViTs in \citet{hu2025arc} that has 10/20 layers, we set 8 layers for the encoder and the decoder, respectively. We set the number of task tokens as 4.

\subsection{Analysis Experiments}

For analysis of attention concentration, we calculate the average Gini metric over all tokens for all heads of the attention of CLS token in the last layer. The Gini metric is computed by:
$$
\text{Gini}(\hat{a}_0) = \frac{\sum_{i=1}^N (2i-N-1)\hat{a}_{0,i}}{\sum_{i=1}^N \hat{a}_{0,i}},
$$
where $\hat{a}_0\in\mathbb{R}^N$ is the sorted attention score ($\hat{a}_{0,1}\leq\hat{a}_{0,2}\leq\hat{a}_{0,N}$) from the CLS token. This metric reflects the unevenness of the attention (essentially concentration measured by paired $L_1$ distance) and is a surrogate for our targeted concentration on relevant tokens, as a larger value typically reflects more concentration but may not focus on all label-relevant parts (which we do not have ground truth). Therefore, this metric may usually first increase and then decrease. Despite this, it can partially reflect the concentration speed and the concentration comparisons between models.

For the synchronization evaluation, we calculate the attention-weighted synchronization of the last layer, i.e.,
$$
\text{Sync}_{\text{att}}(\phi, a_0)= \left\lvert \sum_{j=1}^N a_{0,j} \mathrm{e}^{i\phi_j} \right\rvert=\sqrt{\left(\sum_{j=1}^N a_{0,j}\cos\phi_j\right)^2+\left(\sum_{j=1}^N a_{0,j}\sin\phi_j\right)^2},
$$
where $\phi$ is phases from all tokens while $a_0$ is the attention score from the CLS token. This reflects the group synchronization among attended parts. The results show that this metric continually increases during training, which means that the phases are learned to be synchronized among relevant tokens and also the attention learns to attend on these phase-synchronized parts.

For the visualization of attention maps, we calculate the attention scores for each head of the CLS token of the last layer, where $q$ and $k$ are rotated for KoPE. As KoPE further adopts rotation on $v$ and $o$ for phase interference, we multiply the attention scores by $\lvert\cos(\phi_0-\phi_i))\rvert$ to better reflect the influence of value integration. We present the two best attention heads in the figures.

\section{Detailed Theoretical Analysis}\label{supsec:theory_details}

We consider a simplified theoretical setting for analysis. 
We follow the formulation in \citet{li2023theoretical} and we first introduce the basic settings.

\paragraph{Problem formulation}
We study the binary classification problem that maps $X$ to $y\in\{+1, -1\}$ for any $(X, y)\sim\mathcal{D}$ given $N$ training samples $\left\{(X^n, y^n)_{n=1}^N\right\}$ from the unknown distribution $\mathcal{D}$. Each data point contains $L$ tokens: $X^n=[x_1^n, \cdots, x_L^n]\in\mathbb{R}^{d\times L}$, where each token is unit-norm.

The data is composed of $M$ distinct patterns $\{\mu_1, \mu_2, \cdots, \mu_M\}$, where $\mu_1, \mu_2$ are discriminative patterns while the remaining $M-2$ patterns are non-discriminative. The minimum distance between patterns is denoted by $$\kappa=\min_{1\leq i\neq j\leq M}\lVert\mu_i-\mu_j\rVert>0.$$ Each token is a noisy version of one of the patterns: $\min_{j\in[M]}\lVert x^n_l-\mu_j\rVert\leq\tau$, and the noise is smaller than the separation: $\tau < \kappa/4$.

For each data sample, the label $y^n$ is determined by majority voting of discriminative patterns. The label-relevant tokens, confusion tokens, and label-irrelevant tokens are defined as tokens with the same discriminative pattern as the label, tokens with the different discriminative pattern as the label, and non-discriminative tokens, respectively. The set of relevant tokens is denoted as $S_*$, and $S_{\neg *}$ means other tokens. 
The dataset is assumed balanced, i.e., $\big||\mathcal{D}^+|-|\mathcal{D}^-|\big|=O(\sqrt{N})$, and the average fractions of the label-relevant and confusion tokens over the distribution $\mathcal{D}$ are denoted as $\alpha_*$ and $\alpha_\#$.

\paragraph{Model Training}
In \citet{li2023theoretical}, a simplified shallow ViT with a single-head self-attention layer and a two-layer MLP is considered:
\begin{equation}
    F\!\left(X^{n}\right) = \frac{1}{\left|S^{n}\right|}\sum_{l\in S^{n}}a_{(l)}^{\top}\,\operatorname{ReLU}\!\Bigl(W_{O} W_{V} X^{n}\,\operatorname{softmax}\!\bigl(X^{n\top} W_{K}^{\top} W_{Q} x_{l}^{n}\bigr)\Bigr),
\end{equation}
where $W_{Q}\in\mathbb{R}^{m_{b}\times d}$, $W_{K}\in\mathbb{R}^{m_{b}\times d}$, and $W_{V}\in\mathbb{R}^{m_{a}\times d}$ are weights for query, key, and value vectors, $W_{O}\in\mathbb{R}^{m\times m_{a}}$ is the hidden layer of the MLP (weights for output projection is absorbed in it), $A=(a_{(1)}, a_{(2)}, \cdots, a_{(L)})$ where $a_{(l)}\in\mathbb{R}^{m}$, $l\in [L]$ is the output-layer weights of the MLP, $S^n$ is the set of tokens, and $m_a,m_b>M$. 

Let $\psi=(A,W_{O},W_{V},W_{K},W_{Q})$ denote parameters to train. The training problem is empirical risk minimization: 
$$
    \min_{\psi}:\; f_{N}(\psi)=\frac{1}{N}\sum_{n=1}^{N}\ell(X^{n},y^{n};\psi),
$$
where $\ell(X^{n},y^{n};\psi)$ is the Hinge loss function, i.e.,
$$
    \ell(X^{n},y^{n};\psi)=\max\{1-y^{n}\cdot F(X^{n}),0\}.
$$

The training is solved via mini-batch stochastic gradient descent. $W_V^{(0)}, W_Q^{(0)}, W_K^{(0)}$ comes from an initial model with error $\sigma$, every entry of $W_O$ is generated from $\mathcal{N}(0,\xi^2)$ and every entry of $A$ is sampled from $\left\{+\frac{1}{\sqrt{m}}, -\frac{1}{\sqrt{m}}\right\}$. $A$ is fixed during training.

We slightly modify the model to append a CLS token in the sequence (index 0), while the model output is still defined as the average of all tokens, so the analysis in \citet{li2023theoretical} can be adapted. We mainly focus on the attention from this token.

For KoPE, we consider each token $x$ carries a phase $\phi$. Since a shallow ViT is analyzed, we treat it as the final classification layer of real models, and therefore consider the setting that phases have converged to clustered synchronization. For the rotation in attention, while it can be flexibly learned how to leverage phase differences through projection parameters, we simplify it here as a function $f_a$ depending on the synchronization that intuitively reflects the binding of related tokens (tokens that share a cluster phase reinforce each other in attention and value aggregation; cross-cluster interactions are suppressed), and represent it as a term in the softmax function, i.e., the attention scores are enhanced by $s_{ij} \;:=\; \langle q_i,k_j\rangle + f_a(\cos(\Delta\phi_{ij}))$ where $\Delta\phi_{ij}:=\phi_i-\phi_j$ and $f_a$ is a monotonically increasing function. Let $\Delta\phi_l$ denote the vector of phase difference between token $l$ and the others. Therefore, the model has the formulation:
\begin{equation}
    F\!\left(X^{n}\right) = \frac{1}{\left|S^{n}\right|}\sum_{l\in S^{n}}a_{(l)}^{\top}\,\operatorname{ReLU}\!\Bigl(W_{O} W_{V} X^{n}\,\operatorname{softmax}\!\left(X^{n\top} W_{K}^{\top} W_{Q} x_{l}^{n}+ f_a(\cos(\Delta\phi_l))\right)\Bigr).
\end{equation}

We only modify the logits, leaving $Q,K,V$ parameterization and all training details as in the base analysis. This is only a simplified setting to provide intuitions for binding-like priors, and a conservative use of phase. In real scenarios, phase dynamics and rotations are more complex, and phase differences may be leveraged for more coding purposes and for feature extraction in early layers. We leave more in-depth analysis for future work.

\paragraph{Phase Synchronization} We make some formal characterization of the clustered synchronization of phases.
We assume some tokens are mapped to a cluster for label-relevant parts, while others into different cluster(s). The cluster phase center of the relevant cluster is $\theta_{\mathrm{rel}}$ and the cluster is denoted as $\mathcal{C}$ (others as $\overline{\mathcal{C}}$). We assume the CLS token belongs to the label-relevant cluster through learning or manual specification. For angles we use the distance $\mathrm{dist}(\alpha,\beta):=\min_{m\in\mathbb{Z}}\big|\alpha-\beta+2\pi m\big|\in[0,\pi]$.

\begin{assumption}[Within-cluster tightness]
\label{ass:A1}
For every token $x$ with phase $\phi$ in the cluster $\theta_{\text{rel}}$,
\(
\mathrm{dist}\!\big(\phi,\theta_{\text{rel}}\big)\le \varepsilon_\phi/2,
\)
with a small constant $\varepsilon_\phi\in[0,\pi/2)$. Hence for same-cluster pairs of $\theta_{\text{rel}}$, $\cos(\Delta\phi)\ge \cos(\varepsilon_\phi)$.
\end{assumption}

\begin{assumption}[Inter-cluster separation]
\label{ass:A2}
Inter-cluster separation with other clusters $\Delta\phi_{\text{min}}:=\min_{\phi\in\mathcal{C}, \phi'\in\overline{\mathcal{C}}} \text{dist}(\phi,\phi')$ is larger than within-cluster tightness, so $\rho_\phi=\cos(\Delta\phi_{\text{min}})<\gamma_\phi=\cos(\varepsilon_\phi)$.
\end{assumption}

\begin{assumption}[Cluster--label consistency]
\label{ass:A3}
For each data, the fraction of relevant tokens falling outside $\theta_{\mathrm{rel}}$ cluster is less than a small value $\xi_{\phi}<1-\exp(-(f_a(\gamma_\phi)-f_a(\rho_\phi)))$. No token from other clusters falls in $\theta_{\mathrm{rel}}$ cluster.
\end{assumption}

We show that the phase modulation in attention scores can enlarge the logit gaps for relevant tokens vs. others.
\begin{lemma}
\label{lem:gap}
    Let $S_*$ denote the set of relevant tokens, $\hat{S_*}$ denote the set of relevant tokens that fall in the correct cluster, i.e., $\hat{S_*}:=\{j\in S_*:\pi(x_j)=\text{rel}\}$ ($\hat{S_*}\subseteq S_*, |\hat{S_*}|\ge(1-\xi_\phi)|S_*|$), $\Delta_a := \min_{j\in \hat{S_*}} s_{0j}-\max_{r\in S_{\neg *}} s_{0r}$ denote the logit gap from the CLS token between correct relevant tokens and confusion or irrelevant ones, and $\Delta:=\min_{j\in S_*}\langle q_0,k_j\rangle-\max_{r\in S_{\neg *}}\langle q_0,k_r\rangle$ denote the content-only logit gap between relevant tokens and others. Under Assumptions~\ref{ass:A1}, \ref{ass:A2}, and \ref{ass:A3}, we have $\Delta_a-\Delta\geq f_a(\gamma_\phi)-f_a(\rho_\phi)$.
\end{lemma}
\begin{proof}
    \begin{equation}
        \min_{j\in \hat{S_*}} s_{0j} - \min_{j\in S_*}\langle q_0,k_j\rangle \geq \min_{j\in \hat{S_*}} s_{0j} - \min_{j\in \hat{S_*}}\langle q_0,k_j\rangle \geq \min_{j\in \hat{S_*}}f_a(\cos(\Delta\phi_{0j})) \geq f_a(\gamma_\phi),
    \end{equation}
    so we have
    \begin{equation}
        \Delta_a-\Delta\geq f_a(\gamma_\phi)-\max_{r\in S_{\neg *}} f_a(\cos(\Delta\phi_{0r}))\geq f_a(\gamma_\phi)-f_a(\rho_\phi).
    \end{equation}
\end{proof}

The next lemma states a sufficient condition of the logit gap with correct relevant tokens for $1-\epsilon$ attention concentration on $S_*$.

\begin{lemma}
\label{lem:gap_to_conc}
Let $a_{0j}$ denote the attention weights computed by softmax over logits $s_{0j}$. If
\begin{equation}
\label{eq:gap_thresh}
\Delta_a \ \ge\ \log\frac{|S_{\neg *}|}{(1-\xi_\phi)|S_*|}+\log\frac{1}{\epsilon}=\log\frac{|S_{\neg *}|}{|S_*|}+\log\frac{1}{\epsilon}-\log(1-\xi_\phi),
\end{equation}
then $\sum_{j\in S_*} a_{0j}\ge 1-\epsilon$.
\end{lemma}
\begin{proof}
Let $s_*^{\min}=\min_{j\in \hat{S_*}} s_{0j}$ and $s_{\neg *}^{\max}=\max_{r\in S_{\neg *}} s_{0r}$. Since $\hat{S_*}\subseteq S_*$, then
\begin{equation}
\begin{aligned}
\sum_{j\in S_*} a_{0j}&=
\frac{\sum_{j\in S_*}e^{s_{0j}}}{\sum_{j\in S_*}e^{s_{0j}}+\sum_{r\in S_{\neg *}}e^{s_{0r}}}
\ge
\frac{\sum_{j\in \hat{S_*}}e^{s_{0j}}}{\sum_{j\in \hat{S_*}}e^{s_{0j}}+\sum_{r\in S_{\neg *}}e^{s_{0r}}}
=\frac{1}{1+\frac{\sum_{r\in S_{\neg *}}e^{s_{0r}}}{\sum_{j\in \hat{S_*}}e^{s_{0j}}}}\\
&\ge
\frac{1}{1+\frac{|S_{\neg *}|e^{s_{\neg *}^{\max}}}{|\hat{S_*}|e^{s_*^{\min}}}}
\ge\frac{1}{1+\frac{|S_{\neg *}|}{(1-\xi_\phi)|S_*|}e^{-(s_*^{\min}-s_{\neg *}^{\max})}}.
\end{aligned}
\end{equation}
If \eqref{eq:gap_thresh} holds, then the denominator is at most $1+\epsilon$, hence the fraction is at least $1-\epsilon$.
\end{proof}

From \citet{li2023theoretical}, it has been shown that the dynamics of $W_Q,W_K$ amplify discriminative query/key features, and attention mass concentrates on label-relevant tokens. Since we only modify the logits with an additive term in this setting, the core mechanism of amplification by $W_Q,W_K$ is likely driven by a similar structure of gradients, even accelerated by the structure in the term. Therefore, we assume the underlying amplification dynamics persists under the additive bounded logit perturbation, i.e., the increasing of $\Delta$ (amplifying discriminative query/key features) has similar dynamics.

Then combining the two lemmas yields the relaxation of the condition for attention concentration, likely leading to acceleration.

\begin{lemma}
\label{prop:formal_speedup}
Under Assumptions~\ref{ass:A1}, \ref{ass:A2}, and \ref{ass:A3}, a sufficient condition for $1-\epsilon$ concentration on $S_*$ is
\begin{equation}
\label{eq:shifted_thresh}
\Delta \ \ge\ \log\frac{|S_{\neg *}|}{|S_*|}+\log\frac{1}{\epsilon} - \big(f_a(\gamma_\phi)-f_a(\rho_\phi)+\log(1-\xi_\phi)\big),
\end{equation}
which is relaxed by $f_a(\gamma_\phi)-f_a(\rho_\phi)+\log(1-\xi_\phi)>0$ compared with the baseline threshold $\log\frac{|S_{\neg *}|}{|S_*|}+\log\frac{1}{\epsilon}$. Then under the assumption of a similar training dynamics that increases $\Delta$, attention concentration is achieved earlier.
\end{lemma}

\paragraph{Discussion on training and sample complexity} 
As attention concentration is shown closely related to training and sample complexity~\citep{li2023theoretical}, KoPE may also potentially reduce the complexity. We briefly discuss the parts that the phase modulation can benefit.

\citet{li2023theoretical} derive sufficient condition of training steps and number of training samples for high-probability zero generalization error through a chain: (i) their core SGD dynamics lemma (Lemma 2) establishes that $W_Q,W_K$ amplify discriminative query/key features and induce attention sparsification (Claim 2); (ii) this yields attention concentration (Proposition 2), together with the growth $\|q_1(T)\|^2=\Theta(\log T)$ (Eq.~(63)) and the proxy bound $\phi_n(T)(|S_n|-|S_n^1|)\le \eta^C$ (Eq.~(65)); (iii) these concentration/proxy controls enter the margin lower bound (Eq. (66)) and the subsequent slack term (Eq. (69)), leading to the sufficient iteration requirement (Eq. (70)) and sample bound (Theorem 1). 

KoPE can improve the bottlenecks that feed into training/sample complexity: the residual proxy (Eq.~(65)) and the irrelevant interference terms that are controlled using that residual (Eq.~(66) and Eq.~(69)).

Definition~1 in \citet{li2023theoretical} defines $\phi_n(t),\nu_n(t),p_n(t)$ and Eq.~(65) states
$\phi_n(T)(|S_n|-|S_n^1|)\le \eta^C$ for a large $C$. 
Under our logit shift, the irrelevant-to-relevant exponential ratio is monotonically decreased.

This tightened residual proxy can then improve the iteration and sample requirements. 
In \citet{li2023theoretical}, the sufficient sample bound in Theorem 1 depends on a slack term $c'(1-\zeta)$ (plus $c''(\sigma+\tau)$) introduced in Eq.~(69), which arises from upper bounding non-relevant contributions controlled by attention concentration/proxy quantities (Eq.~(65)–(66)). Under our additive logit shift, we have a relaxed sufficient condition for attention concentration from the CLS query, hence the non-relevant exponential mass entering the proxy bound is reduced at the same stage of training. This suggests a smaller effective constant in the slack term $c'(1-\zeta)$, and therefore potentially smaller sufficient requirements on iterations and samples, consistent with the observed training/data efficiency.

\section{More Experiment Results}\label{supsec:experiment_results}

\subsection{Longer Supervised Training}\label{supsec:longer}

\begin{table}[tb]
  \centering
  \caption{Supervised learning results on ImageNet-1K with different training epochs.}
  \label{suptab:longer}
  \setlength{\tabcolsep}{10pt}

  \begin{tabular}{c l c c c c}
    \toprule
    \multirow{2}{*}{\textbf{Scale}} &
    \multirow{2}{*}{\textbf{Model}} &
    \multicolumn{2}{c}{\textbf{300 epochs}} & 
    \multicolumn{2}{c}{\textbf{800 epochs}} \\
    \cmidrule(lr){3-4} \cmidrule(lr){5-6}
    & & \textbf{val} & \textbf{V2} & \textbf{val} & \textbf{V2} \\
    \midrule

    \multirow{2}{*}{small} &
    ViT & 80.1 & 68.7 & 81.3 & 70.3 \\
    & ViT+KoPE & \textbf{80.8} & \textbf{69.9} & \textbf{82.1} & \textbf{71.4} \\
    \midrule

    \multirow{2}{*}{base} &
    ViT & 82.1 & 71.2 & 82.9 & 72.5 \\
    & ViT+KoPE & \textbf{83.0} & \textbf{72.7} & \textbf{83.2} & \textbf{73.2} \\
    \midrule

    \multirow{2}{*}{large} &
    ViT & 82.8 & 71.8 & 83.6 & 73.2 \\
    & ViT+KoPE & \textbf{84.0} & \textbf{73.8} & \textbf{84.1} & \textbf{74.4} \\
    \bottomrule
  \end{tabular}
\end{table}

As our main experiments train models for 300 epochs, we further evaluate longer training time to demonstrate that the advantage of KoPE persists. As shown in Table~\ref{suptab:longer}, the performance improvement of KoPE over vanilla ViT remains under longer training, especially for the generalization on ImageNet V2. Meanwhile, the results also show that ViT+KoPE trained by 300 epochs can achieve similar or better performance than vanilla ViT trained by 800 epochs, further demonstrating the learning efficiency of KoPE.

\subsection{Experiments with Matched FLOPs}\label{supsec:match-flops}

\begin{table}[t]
  \centering
  \caption{Supervised learning dynamics of ViT + KoPE with matched FLOPs as ViT (non-EMA models). ``ep'' refers to the epoch.}
  \label{suptab:match-flops}
  \begin{tabular}{l|c|c|rrrrr}
    \toprule
    Model & Parameters & FLOPs & 25 ep & 50 ep & 100 ep & 200 ep & 300 ep \\
    \midrule
    ViT & 86.6M & 17.6G & 33.64 & 57.15 & 70.78 & 80.06 & 81.98 \\
    ViT + KoPE (original) & 87.9M & 21.1G & 48.14 & 63.83 & 73.36 & 80.92 & 82.85 \\
    ViT + KoPE (MLP ratio = 2.75) & 70.2M & 17.6G & 48.36 & 64.00 & 73.17 & 80.82 & 82.82 \\
    ViT + KoPE (width = 704) & 74.0M & 17.8G & 47.38 & 63.26 & 72.85 & 80.73 & 82.70 \\
    \bottomrule
  \end{tabular}
\end{table}

To further exclude the influence of additional computational capacity, we conduct experiments to reduce the FLOPs of ViT+KoPE-B to approximately match ViT-B, while having fewer parameters. We either (i) reduce the MLP ratio from 4 to 2.75, or (ii) reduce the network width from 768 to 704 (with 11 attention heads).

As shown in Table~\ref{suptab:match-flops}, ViT+KoPE remains superior under similar FLOPs and fewer parameters, indicating that gains stem from the synchronization mechanism rather than additional computational capacity.

\subsection{More Ablation Analysis}\label{supsec:ablation}

\begin{figure}
    \centering
    \begin{subfigure}[t]{0.4\textwidth}
      \centering
      \includegraphics[width=\linewidth]{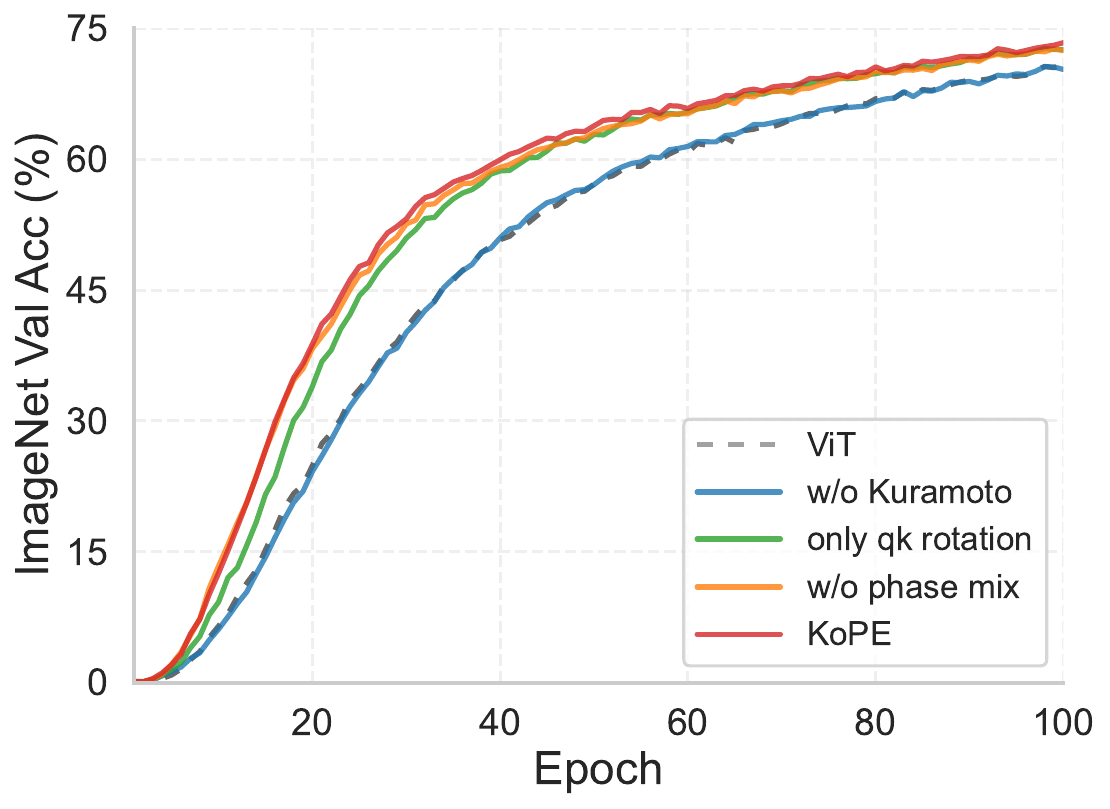}
      \subcaption{Early stage}
    \end{subfigure}
    \begin{subfigure}[t]{0.4\textwidth}
      \centering
      \includegraphics[width=\linewidth]{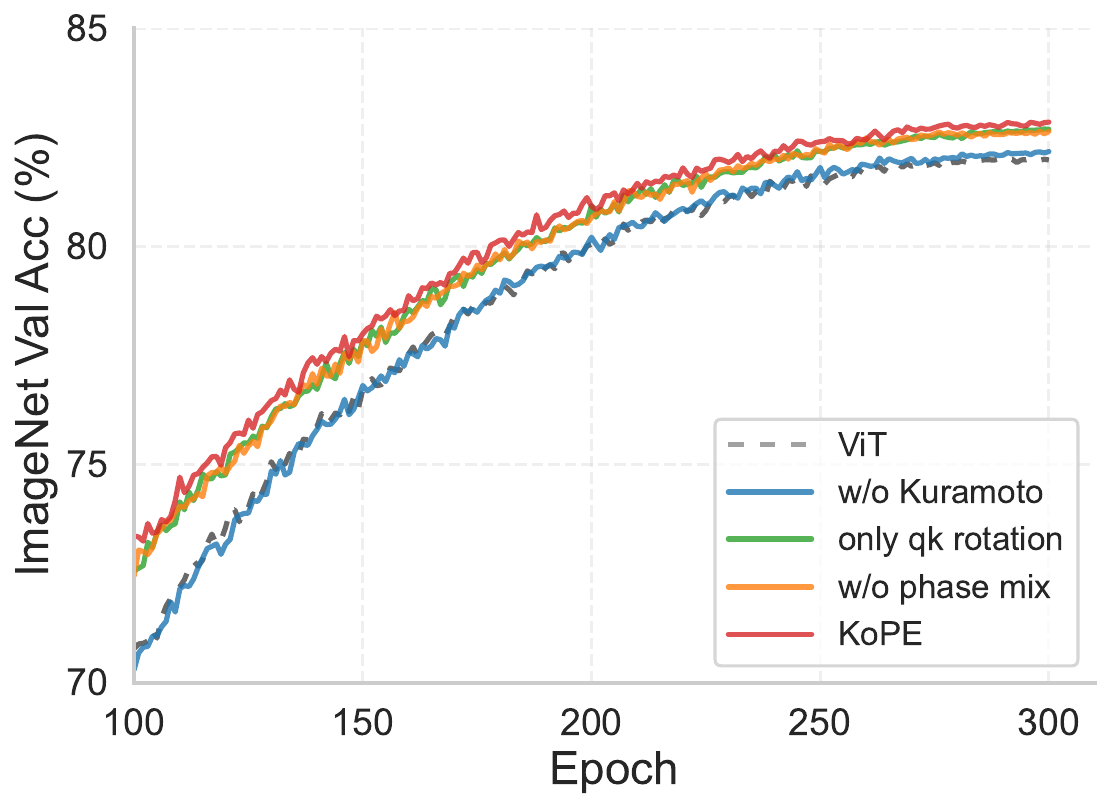}
      \subcaption{Late stage}
    \end{subfigure}
    \caption{More ablation analyses. ``w/o Kuramoto'' uses fixed phase states as initialization. ``only qk rotation'' applies phase rotation only to query and key vectors. ``w/o phase mix'' does not adopt the mixture of phases.}
    \label{supfig:ablation}
\end{figure}

\begin{table}
  \centering
  \caption{Supervised learning dynamics of the baseline that learns phases from token representations. ``ep'' refers to the epoch.}
  \label{suptab:phase-from-token}
  \begin{tabular}{l|rrrrr}
    \toprule
    Model & 25 ep & 50 ep & 100 ep & 200 ep & 300 ep \\
    \midrule
    Phase learned from token & 25.80 & 46.64 & 65.57 & 77.69 & 80.72 \\
    KoPE & 48.14 & 63.83 & 73.36 & 80.92 & 82.85 \\
    \bottomrule
  \end{tabular}
\end{table}

\begin{table}
  \centering
  \caption{Supervised learning dynamics of the baselines with lower softmax temperatures in attention. ``ep'' refers to the epoch.}
  \label{suptab:temperature-tuning}
  \begin{tabular}{l|rrrrr}
    \toprule
    Model & 25 ep & 50 ep & 100 ep & 200 ep & 300 ep \\
    \midrule
    ViT & 33.64 & 57.15 & 70.78 & 80.06 & 81.98 \\
    ViT (temp=0.7) & 33.76 & 57.20 & 70.00 & 79.91 & 82.00 \\
    ViT (temp=0.5) & 35.32 & 57.55 & 70.06 & 79.79 & 81.81 \\
    ViT + KoPE & 48.14 & 63.83 & 73.36 & 80.92 & 82.85 \\
    \bottomrule
  \end{tabular}
\end{table}

In Fig.~\ref{supfig:ablation}, we further present more analysis of the components in our implementation. The results show that value and output rotation has efficiency gains at the early stage over only query and key rotation in the attention module, while the final performance is similar. Additionally, the mixture of phases has slight final performance gain. The major improvement comes from the Kuramoto dynamics.

We also investigate a baseline: replacing Kuramoto dynamics with a shared learnable module that learns to derive (relational) phases from token representations at each layer. As shown in Table~\ref{suptab:phase-from-token}, this baseline is significantly weaker throughout training, suggesting that the gains are mainly from the synchronization dynamics prior rather than a learned relational bias.

To further verify that the gains are from the data-specific attention concentration via KoPE rather than generally sparser attention, we compare KoPE with two ViT baselines with lower softmax temperatures (0.7 or 0.5) that encourage sparser attention. As shown in Table~\ref{suptab:temperature-tuning}, temperature tuning of ViT does not yield gains as KoPE, demonstrating the unique advantage of KoPE.

\subsection{Training Dynamics on Image Segmentation}\label{supsec:dynamics-segmentation}

\begin{figure}[t]
    \centering
    \begin{subfigure}[t]{0.32\textwidth}
      \centering
      \includegraphics[width=\linewidth]{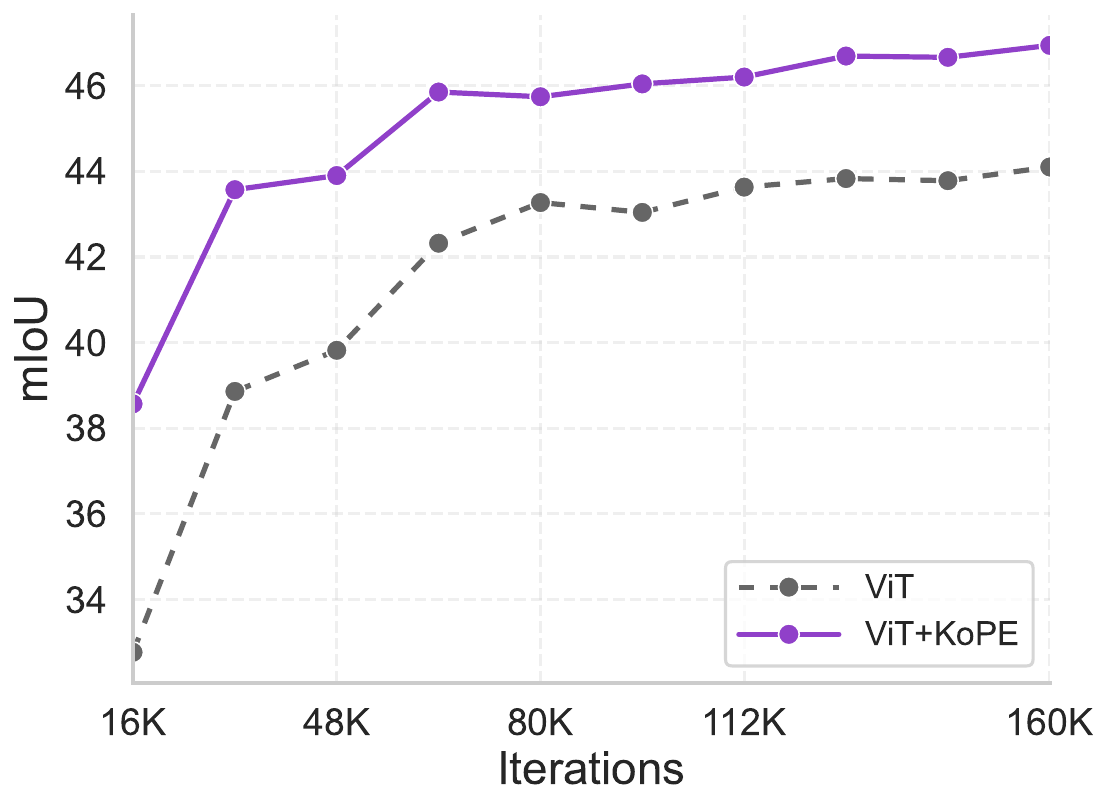}
      \subcaption{ADE-20K, base model}
    \end{subfigure}
    \begin{subfigure}[t]{0.32\textwidth}
      \centering
      \includegraphics[width=\linewidth]{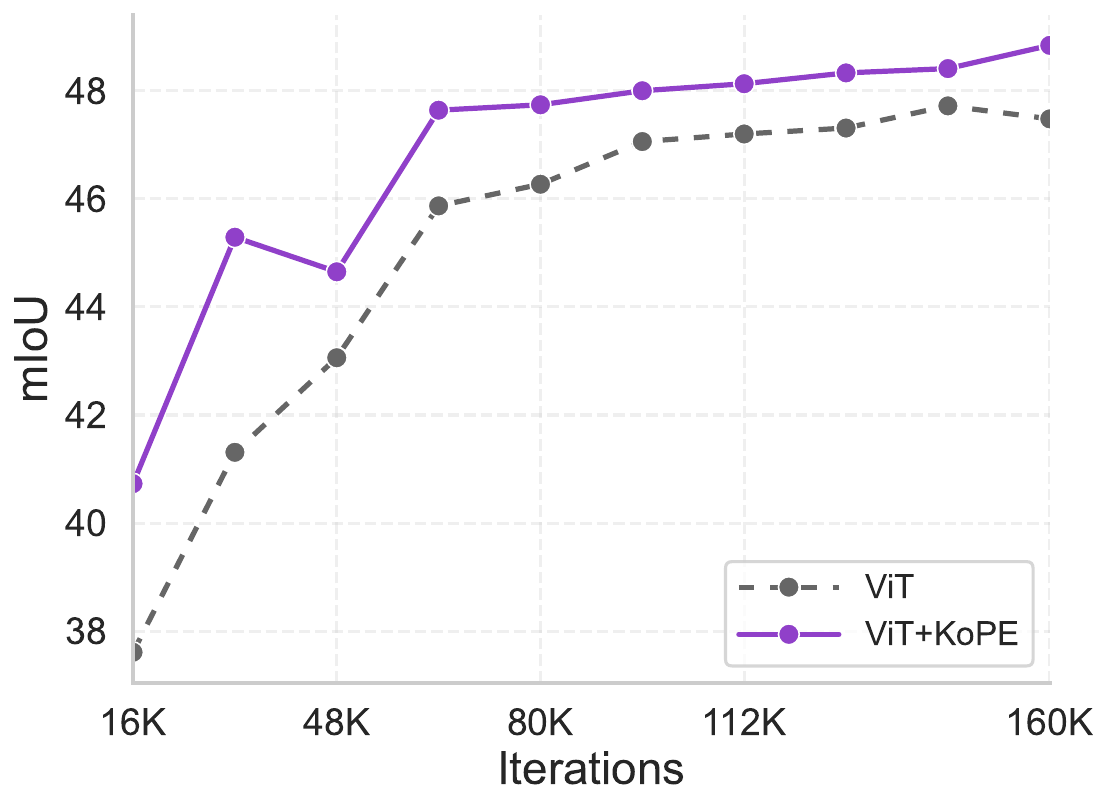}
      \subcaption{ADE-20K, large model}
    \end{subfigure}
    \begin{subfigure}[t]{0.32\textwidth}
      \centering
      \includegraphics[width=\linewidth]{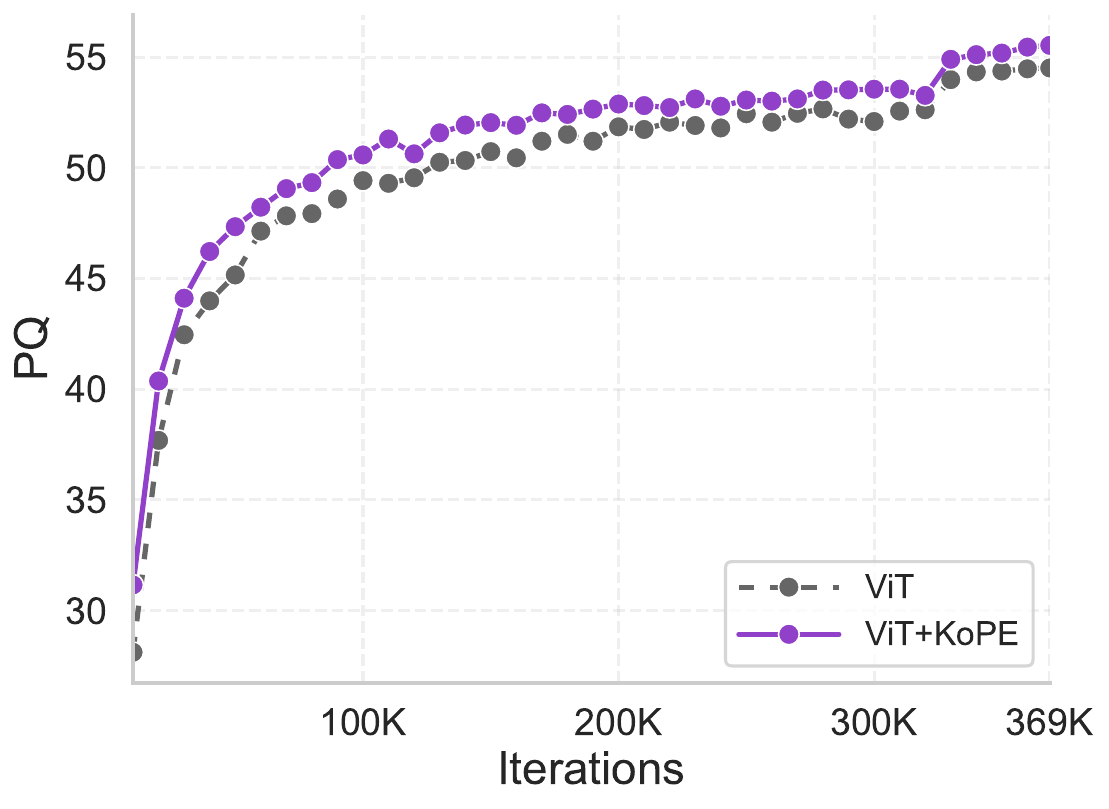}
      \subcaption{COCO}
    \end{subfigure}
    \caption{Training dynamics on image segmentation tasks (ADE-20K semantic segmentation with SETR-PUP framework, COCO panoptic segmentation with Mask2Former framework).}
    \label{supfig:segmentation_training}
\end{figure}

In Fig.~\ref{supfig:segmentation_training}, we present the training dynamics of ViT+KoPE and ViT for image segmentation tasks. It shows that KoPE also demonstrates large training efficiency, especially on ADE-20K with SETR-PUP that better reflects the ability of ViT-style backbones (achieving better performance with less than 40\% of training time). For Mask2Former with heavy segmentation heads, ViT+KoPE backbone still enhances the learning efficiency and final performance. Future work can further explore integrating KoPE into modules in these segmentation heads.

\begin{figure}[t]
    \centering
    \begin{subfigure}[t]{0.19\textwidth}
      \centering
      \includegraphics[width=\linewidth]{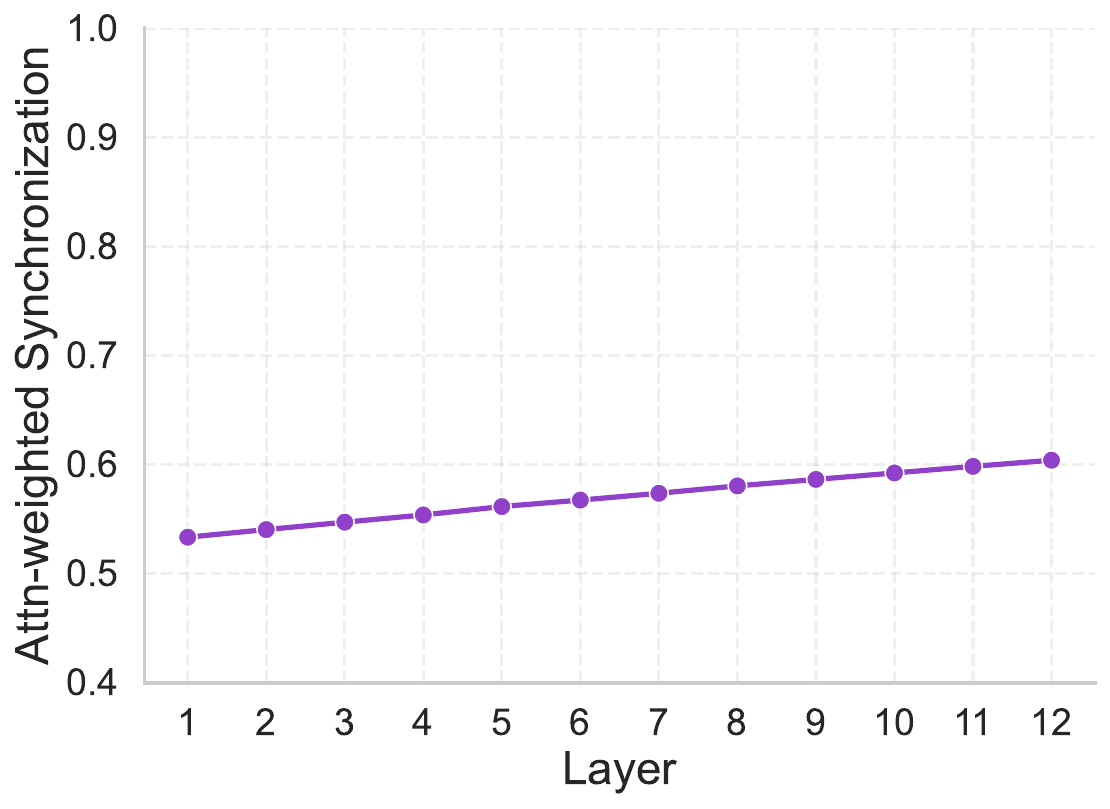}
      \subcaption{Epoch 1}
    \end{subfigure}
    \begin{subfigure}[t]{0.19\textwidth}
      \centering
      \includegraphics[width=\linewidth]{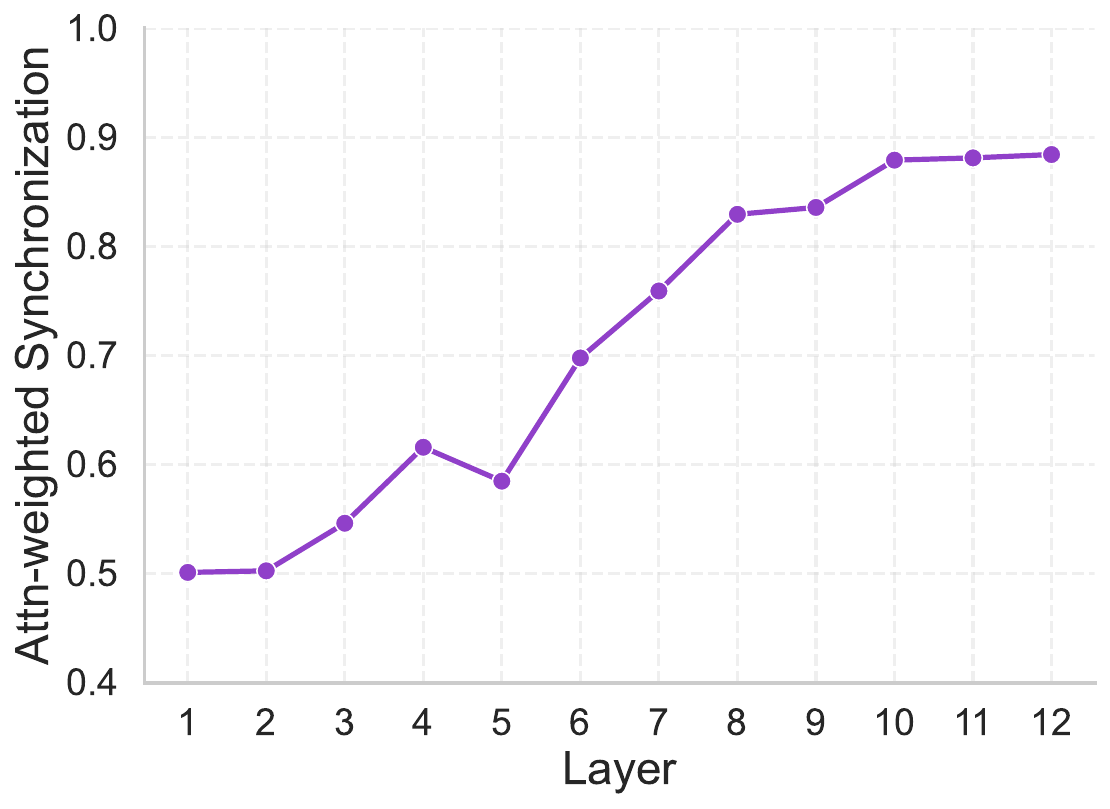}
      \subcaption{Epoch 50}
    \end{subfigure}
    \begin{subfigure}[t]{0.19\textwidth}
      \centering
      \includegraphics[width=\linewidth]{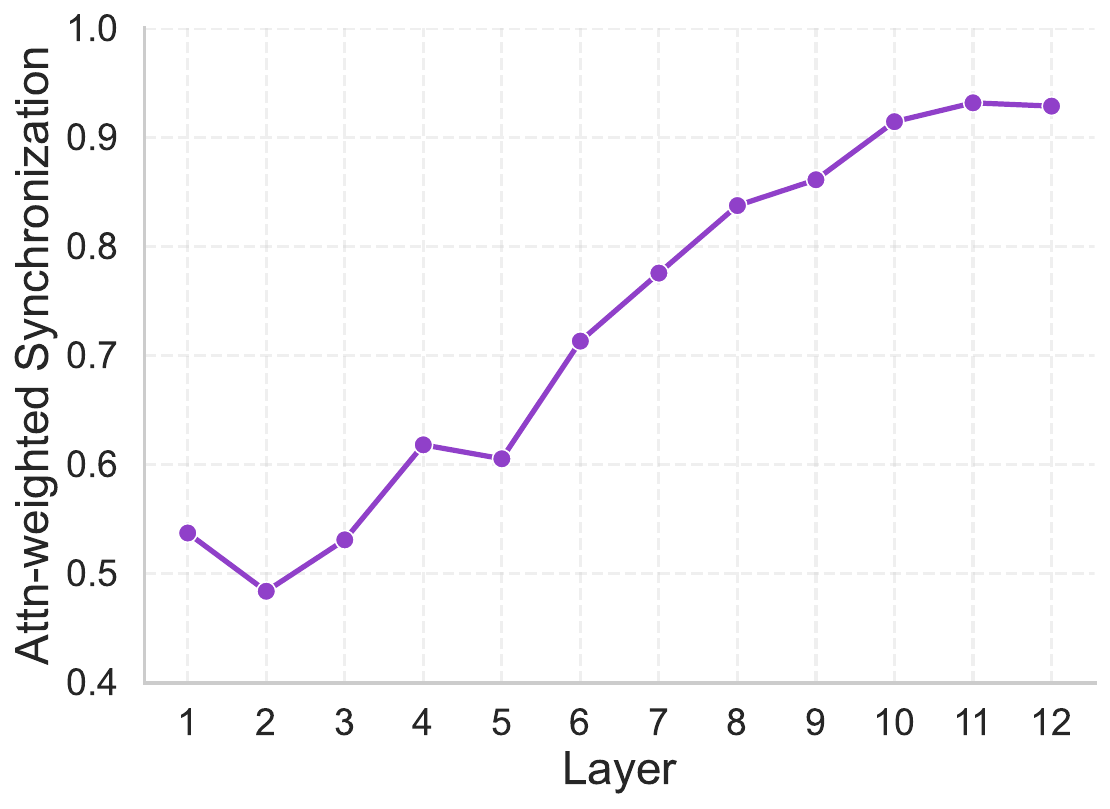}
      \subcaption{Epoch 100}
    \end{subfigure}
    \begin{subfigure}[t]{0.19\textwidth}
      \centering
      \includegraphics[width=\linewidth]{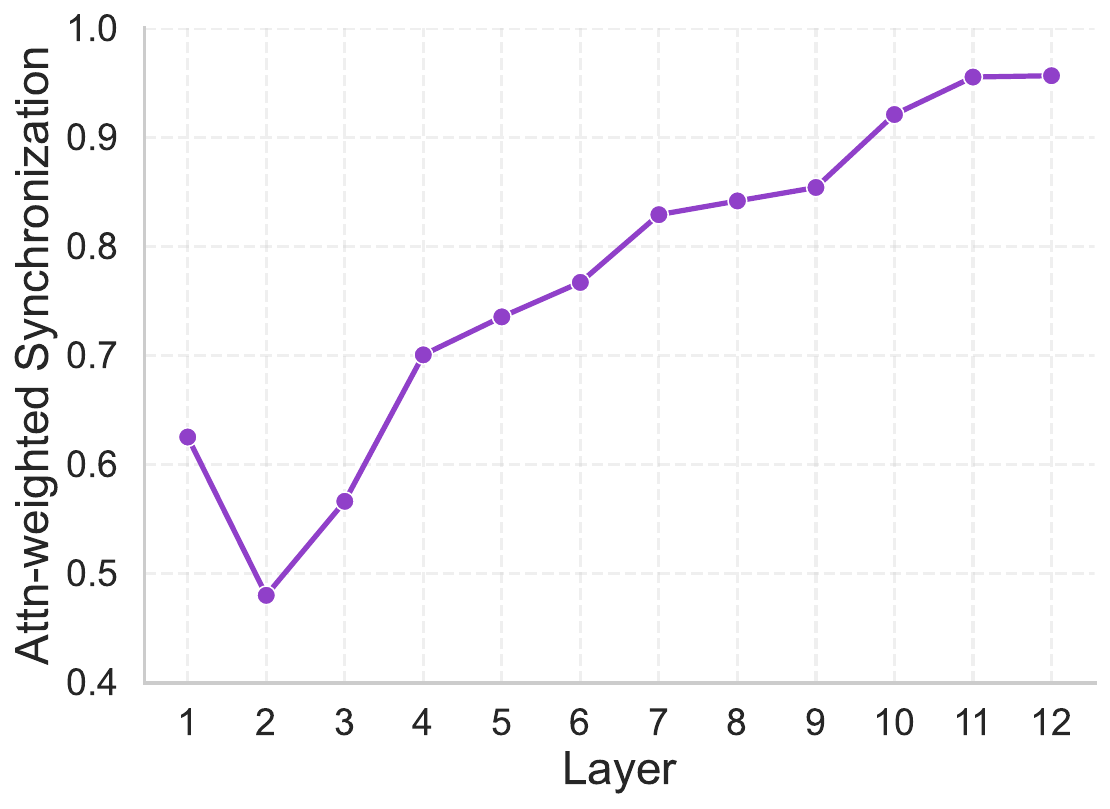}
      \subcaption{Epoch 200}
    \end{subfigure}
    \begin{subfigure}[t]{0.19\textwidth}
      \centering
      \includegraphics[width=\linewidth]{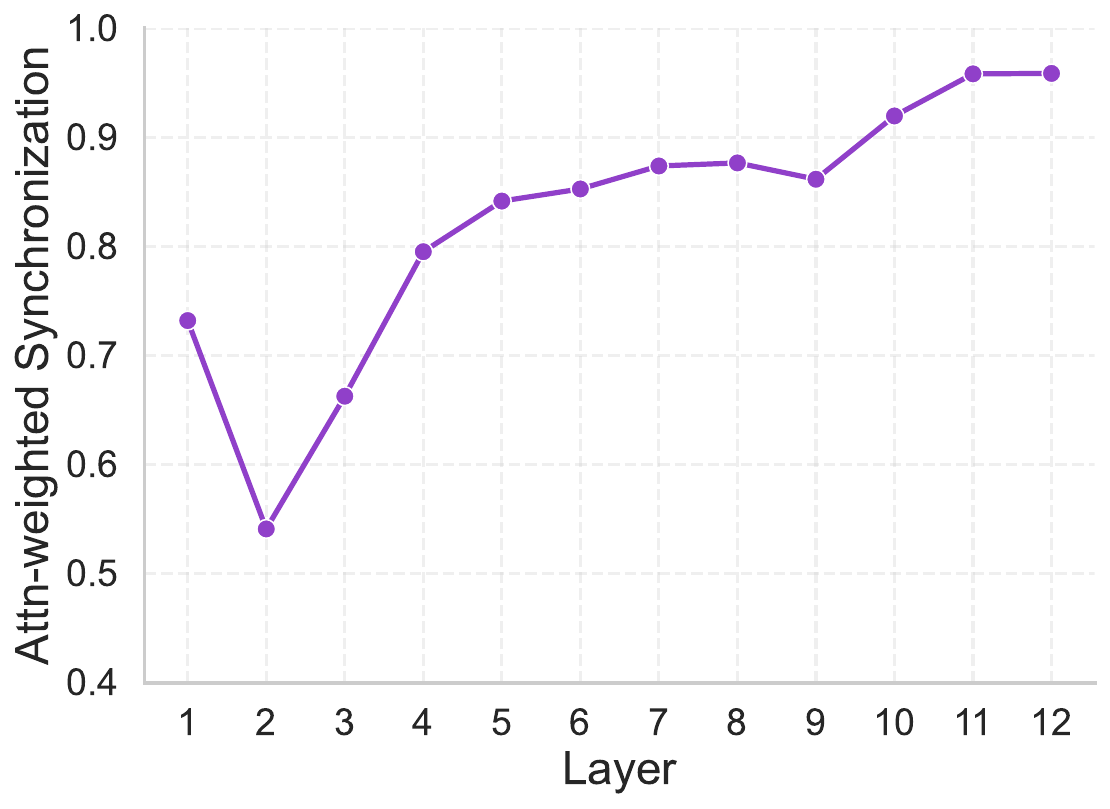}
      \subcaption{Epoch 300}
    \end{subfigure}
    \caption{Comparison of attention-weighted phase synchronization through layers during training.}
    \label{supfig:phase_sync}
\end{figure}

\subsection{More Analysis Results}\label{supsec:more-analysis}

\paragraph{Phase synchronization}
In Fig.~\ref{supfig:phase_sync}, we demonstrate the attention-weighted phase synchronization through layers during different training stages. Initially, attention and coupling parameters are random so the synchronization only slightly increases along layers due to the prior of Kuramoto dynamics. During training, KoPE learns to synchronize related parts and the attention module learns to focus on relevant tokens, leading to increased (attended) group synchronization at later layers.

\begin{table}[h!]
  \centering
  \caption{Supervised learning dynamics of different coupling types. ``ep'' refers to the epoch.}
  \label{suptab:coupling-type}
  \begin{tabular}{l|rrrrr}
    \toprule
    Coupling Type & 25 ep & 50 ep & 100 ep & 200 ep & 300 ep \\
    \midrule
    Cosine similarity & 47.33 & 63.16 & 73.06 & 80.60 & 82.75 \\
    Softmax & 48.14 & 63.83 & 73.36 & 80.92 & 82.85 \\
    \bottomrule
  \end{tabular}
\end{table}

\begin{table}[t]
  \centering
  \caption{Supervised learning dynamics of different $\gamma$. ``ep'' refers to the epoch.}
  \label{suptab:analyze-gamma}
  \begin{tabular}{l|rrrrr}
    \toprule
    $\gamma$ & 25 ep & 50 ep & 100 ep & 200 ep & 300 ep \\
    \midrule
    0.01 & 48.88 & 64.01 & 73.67 & 80.93 & 82.81 \\
    0.05 (in main text) & 48.14 & 63.83 & 73.36 & 80.92 & 82.85 \\
    learnable, init 0.05 & 48.59 & 63.93 & 73.46 & 81.03 & 82.76 \\
    0.1 & 48.96 & 63.88 & 73.44 & 80.91 & 82.92 \\
    0.2 & 48.64 & 64.04 & 73.28 & 80.97 & 82.94 \\
    0.5 & 47.09 & 63.96 & 73.60 & 80.93 & 82.66 \\
    1.0 & 46.34 & 63.26 & 73.24 & 80.82 & 82.68 \\
    \bottomrule
  \end{tabular}
\end{table}

\begin{table}[t]
  \centering
  \caption{Supervised learning dynamics of different phase initialization. ``ep'' refers to the epoch.}
  \label{suptab:analyze-phase-init}
  \begin{tabular}{l|rrrrr}
    \toprule
    Initialization type & 25 ep & 50 ep & 100 ep & 200 ep & 300 ep \\
    \midrule
    ViT & 33.64 & 57.15 & 70.78 & 80.06 & 81.98 \\
    ViT+KoPE, random init & 26.50 & 45.91 & 62.49 & 76.12 & 79.40 \\
    ViT+KoPE, learned init & 36.05 & 59.66 & 72.41 & 80.58 & 82.87 \\
    ViT+KoPE, 2D rotary init & 48.14 & 63.83 & 73.36 & 80.92 & 82.85 \\
    \bottomrule
  \end{tabular}
\end{table}

\begin{table}[h!]
  \centering
  \caption{Semantic segmentation results (mIoU) on subsets of ADE20K (val) with SETR-PUP under single-scale evalution.}
  \label{suptab:ade20k-subset}
  \begin{tabular}{lcc}
    \toprule
    Split & ViT-B & ViT+KoPE-B \\
    \midrule
    Full & 44.1 & \textbf{46.9 (+2.8)} \\
    Complex subset & 34.4 & \textbf{39.3 (+4.9)} \\
    \bottomrule
  \end{tabular}
\end{table}

We also analyze the phase synchronization on ADE20K and COCO, using the entity labels available in each dataset (semantic-class labels for ADE20K and instance labels for COCO). We compute an average phase synchronization score among tokens belonging to the same entity over the test dataset. Specifically, for an entity $e$ with token set $T_e$, where each token has $P$ phases $\phi$ at layer $l$, we compute a phase synchronization score:
$$\text{Sync}_l^e=\frac{1}{P}\sum_{p=1}^P \left|\frac{1}{|T_e|}\sum_{j\in T_e} \text{e}^{i\phi_{j, l}^{(p)}}\right|,$$
which measures the group synchronization of phases for this entity. We then average it over all entities across the whole dataset. 
Since we have multiple phases for each token and employ a phase-mixing module, we report the average score for both raw and mixed phases. Across depth, synchronization increases, especially after mixing:
\begin{itemize}
    \item ADE20K (raw): layer 0 = 0.597 → layer 11 = 0.757; (mix): 0.771 → 0.963;
    \item COCO (raw): layer 0 = 0.543 → layer 11 = 0.669; (mix): 0.677 → 0.936.
\end{itemize}
These statistics provide quantitative evidence that phase states become more aligned for same-entity tokens across layers.

\paragraph{Coupling type}
While we choose attention-like softmax computation for coupling strength computation, it is also possible for other types, e.g., cosine similarity that allows for both positive and negative interactions. We conduct experiments to replace the softmax computation by cosine similarity between $h_q(\mathbf{z}^l)$ and $h_k(\mathbf{z}^l)$ (divided by $\sqrt{d}$). As shown in Table~\ref{suptab:coupling-type}, the results are similar, indicating that KoPE is relatively robust to the coupling type.

\paragraph{Sensitivity of $\gamma$}

We further study the sensitivity of the hyper-parameter $\gamma$. We sweep fixed $\gamma$ from 0.01 to 1.0, and also test the learnable $\gamma$ with initialization 0.05. As shown in Table~\ref{suptab:analyze-gamma}, the learning curves and final performance are similar, indicating that KoPE is relatively robust to $\gamma$ within certain range, as the model may learn to adapt to different step size.

\paragraph{Phase initialization}

We also investigate the importance of a meaningful initialization for phases, e.g., 2D rotary initialization. We test alternative initializations: fixed random phases or learned initialization (phase predicted from initial token representations by a learnable module). As shown in Table~\ref{suptab:analyze-phase-init}, random init hurts substantially, as it confuses the early layers with meaningless phases, while learned init has similar final performance but slower learning speed. This suggests the importance of a meaningful phase initialization.

\paragraph{Complex subset of ADE20K}

To further demonstrate the advantage of KoPE for structured understanding, we analyze a complex subset of ADE20K: images with $\geq 15$ distinct semantic classes and $\geq 6$ connected components with area below 1\% of the image, which constitute around 10\% of the original test data. This represents a challenging real-world scenario where structural binding for multiple (small) components is important. As shown in Table~\ref{suptab:ade20k-subset}, KoPE's advantage is amplified on this harder slice, indicating the potential of KoPE for complex real-world scenarios.

\subsection{Complete CLIP Benchmark Results}\label{supsec:clip}

\begin{table}[h]
    \centering
    \caption{Summary results of comparison on 40 tasks in CLIP benchmark. ``T2I'' and ``I2T'' mean text-to-image and image-to-text zero-shot retrieval.}
    \label{suptab:clip}
    \setlength{\tabcolsep}{8pt}
    {\textbf{Zero-shot Classification (Acc@1)}\par}
    \begin{tabular}{llccc}
    \toprule
    Type & Dataset & ViT-B & ViT+KoPE-B & $\Delta$ \\
    \midrule
    \multirow{7}{*}{\makecell[l]{ImageNet variants\\ \& ObjectNet}} &
    ImageNet-1K     & 28.21 & 29.22 & $+$1.01 \\
    & ImageNet-A      & 6.45  & 6.76  & $+$0.31 \\
    & ImageNet-O      & 39.50 & 37.85 & $-$1.65 \\
    & ImageNet-R      & 28.93 & 32.30 & $+$3.36 \\
    & ImageNet-Sketch & 16.38 & 18.86 & $+$2.48 \\
    & ImageNet-V2     & 22.77 & 24.21 & $+$1.44 \\
    & ObjectNet       & 26.62 & 29.31 & $+$2.69 \\
    \midrule
    \multirow{10}{*}{\makecell[l]{Others}} &
    Cars          & 28.29 & 32.40 & $+$4.10 \\
    & Country211    & 6.06  & 6.30  & $+$0.25 \\
    & FER2013       & 18.14 & 15.46 & $-$2.67 \\
    & FGVC-Aircraft & 1.77  & 1.92  & $+$0.15 \\
    & GTSRB         & 12.06 & 13.33 & $+$1.27 \\
    & MNIST         & 5.51  & 12.04 & $+$6.53 \\
    & RenderedSST2  & 48.98 & 49.75 & $+$0.77 \\
    & STL10         & 83.91 & 86.44 & $+$2.53 \\
    & SUN397        & 41.14 & 42.33 & $+$1.19 \\
    & VOC2007       & 61.14 & 66.51 & $+$5.36 \\
    \midrule
    \multirow{20}{*}{\makecell[l]{Visual Task\\ Adaptation Benchmark}} &
    Caltech101                      & 70.83 & 72.70 & $+$1.87 \\
    & CIFAR-10                        & 80.03 & 80.45 & $+$0.42 \\
    & CIFAR-100                       & 42.02 & 49.39 & $+$7.37 \\
    & CLEVR closest object distance   & 15.79 & 21.92 & $+$6.13 \\
    & CLEVR count all                 & 12.76 & 13.01 & $+$0.25 \\
    & Diabetic Retinopathy            & 7.92  & 60.44 & $+$52.52 \\
    & DMLab                           & 14.86 & 14.89 & $+$0.04 \\
    & dSprites label orientation      & 2.63  & 2.99  & $+$0.36 \\
    & dSprites label x position       & 3.14  & 2.48  & $-$0.65 \\
    & dSprites label y position       & 3.28  & 3.26  & $-$0.02 \\
    & DTD                            & 23.40 & 24.52 & $+$1.12 \\
    & EuroSAT                        & 32.43 & 32.41 & $-$0.02 \\
    & flowers                        & 16.64 & 18.39 & $+$1.76 \\
    & KITTI closest vehicle distance & 27.29 & 29.11 & $+$1.83 \\
    & PCam                           & 58.20 & 59.77 & $+$1.57 \\
    & pets                           & 28.73 & 28.13 & $-$0.60 \\
    & RESISC45                       & 24.87 & 22.35 & $-$2.52 \\
    & SmallNORB label azimuth        & 4.91  & 5.42  & $+$0.51 \\
    & SmallNORB label elevation      & 12.49 & 12.03 & $-$0.45 \\
    & SVHN                           & 7.77  & 11.32 & $+$3.55 \\
    \bottomrule
    \end{tabular}
    \par\vspace{4mm}
    {\textbf{Zero-shot Retrieval (Recall@5)}\par}
    \begin{tabular}{l|ccc|ccc}
    \toprule
    Dataset & T2I (ViT) & T2I (ViT+KoPE) & $\Delta$ & I2T (ViT) & I2T (ViT+KoPE) & $\Delta$ \\
    \midrule
    Flickr30k & 49.56 & 50.32 & $+$0.76  & 61.90 & 63.40 & $+$1.50 \\
    Flickr8k  & 45.72 & 45.58 & $-$0.14 & 59.30 & 58.80 & $-$0.50 \\
    MSCOCO    & 29.56 & 31.45 & $+$1.88  & 42.56 & 44.54 & $+$1.98 \\
    \bottomrule
    \end{tabular}
\end{table}

In Table~\ref{suptab:clip}, we present the complete results of ViT and ViT+KoPE after vision-language learning on 40 tasks in CLIP benchmark. The results show that KoPE has better performance on most datasets and on average, demonstrating that it encourages structured representation learning that can be better aligned with symbolic languages and has generally better generalization performance.

We note a large variability of performance on the Diabetic Retinopathy. This phenomenon has also been reported in the literature~\citep{schuhmann2022laion}, where they find accuracy goes from 3\% to 73.3\% for different models. This is likely because the dataset is highly imbalanced and the evaluation may be sensitive to the prompt. Excluding this subset does not change our main conclusion.

\begin{figure}
    \centering
    \includegraphics[width=\linewidth]{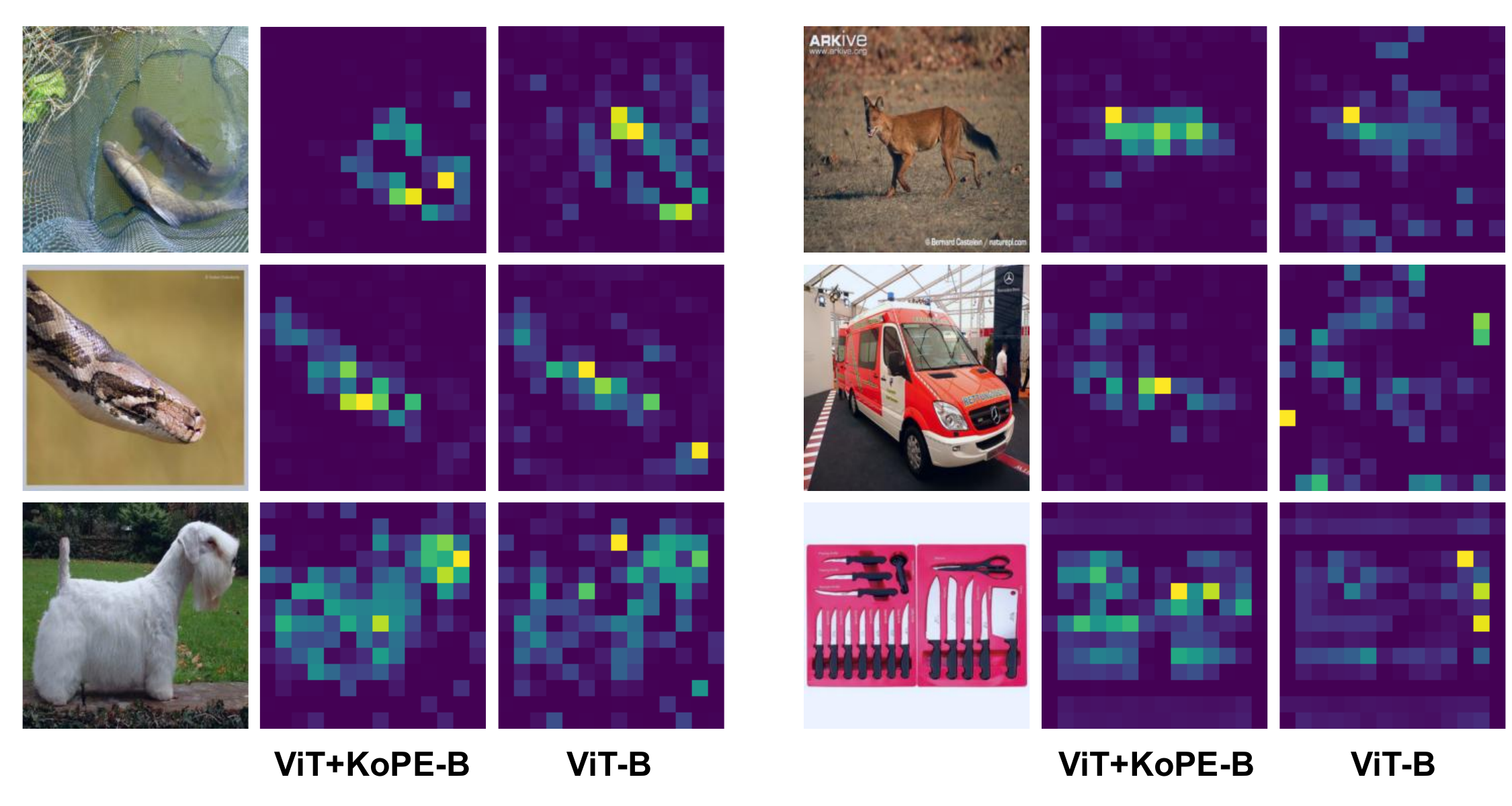}
    \caption{More visualization of attention maps from the last layer's CLS token for ViT and ViT+KoPE under supervised training on ImageNet-1K. KoPE facilitates attention concentration on relevant tokens even under supervised learning, which was believed to fall short in this objective.}
    \label{supfig:visualization1}
\end{figure}

\begin{figure}
    \centering
    \includegraphics[width=\linewidth]{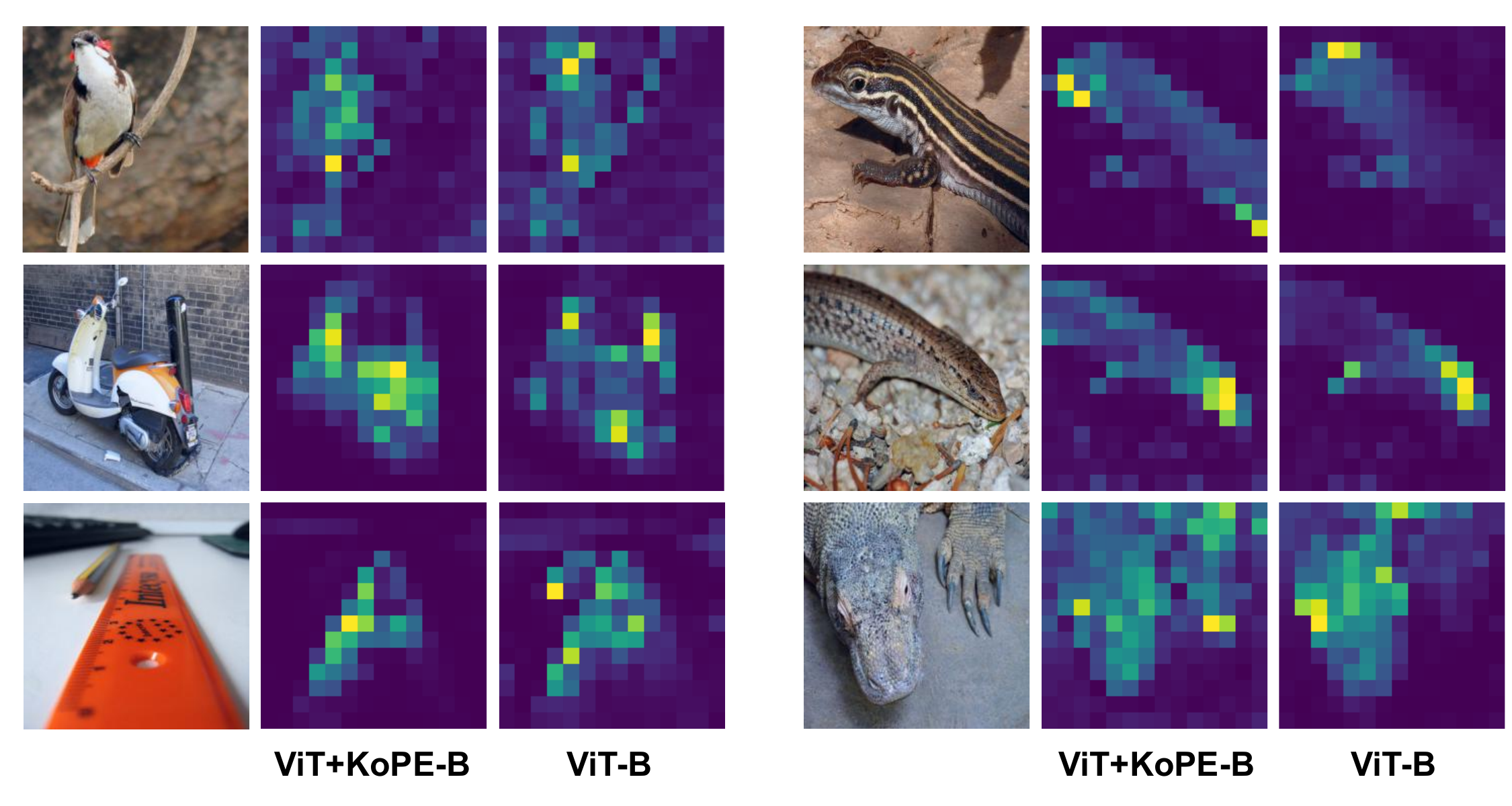}
    \caption{More visualization of attention maps from the last layer's CLS token for ViT and ViT+KoPE under self-supervised learning (SimDINOv2) on ImageNet-1K. The left images show that compared with vanilla ViT, KoPE reduces attention to non-/different-object parts, while the right images demonstrate that KoPE encourages binding of the whole entities. The results indicate that KoPE can further advance self-supervised learning that has been shown to be able to induce emergent object representations.}
    \label{supfig:visualization2}
\end{figure}

\subsection{More Visualization}\label{supsec:visualization}

In Fig.~\ref{supfig:visualization1}, we provide more visualization of attention maps of the best head from the last layer's CLS token for ViT and ViT+KoPE under supervised training. The results show that KoPE facilitates attention concentration on relevant tokens, forming more structured representations even under supervised learning that was believed to fall short in object representations~\citep{caron2021emerging}.

Additionally, we provide more visualization of attention maps under self-supervised learning~\citep{wusimplifying}, which has been shown to benefit object representations~\citep{caron2021emerging}. As shown in Fig.~\ref{supfig:visualization2}, KoPE can further reduce attention to non-object or different-object parts or facilitate the binded attention to the whole entities, compared to ViT-B trained by SimDINOv2. The results demonstrate the complementary advantages of KoPE over learning paradigms.

\section{More Discussions}\label{supsec:discussion}

Spatiotemporal dynamics are important information processing mechanisms in neuroscience. It has also been investigated in the area of spiking neural networks~\citep{maass1997networks}, where spiking time can be used for efficient coding~\citep{mostafa2017supervised}, relational processing~\citep{xiao2024temporal}, or adversarial robustness~\citep{ding2025neuromorphic}. Nevertheless, due to the discrete spikes and unfolded time steps, they are hard to scale up with rich spatiotemporal properties. Differently, this paper considers Kuramoto dynamics of phases for temporal information with synchronization and injects it into the depth evolution of mainstream artificial neural networks. This abstraction of neuro-inspired mechanisms enables scalability to advance state-of-the-art deep learning models.

We mainly focus on vision tasks in this paper as they are closely related to structured understanding from unstructured data through binding-like integration. We show that synchronization can serve as a scalable neuro-inspired mechanism, and it is possible for future extension to other modalities, as well as multimodal integration requiring binding among different modalities to formulate coherent representations for concepts or entities.

This paper mainly investigates neural architectures from the perspective of neural representation and information processing, which is compatible with mainstream connectivity design. Future works can further explore the combination of efficient connectivity design, e.g., efficient attention/interactive modules, with the components in KoPE to further advance the efficiency of the model.

%%%%%%%%%%%%%%%%%%%%%%%%%%%%%%%%%%%%%%%%%%%%%%%%%%%%%%%%%%%%%%%%%%%%%%%%%%%%%%%
%%%%%%%%%%%%%%%%%%%%%%%%%%%%%%%%%%%%%%%%%%%%%%%%%%%%%%%%%%%%%%%%%%%%%%%%%%%%%%%

\end{document}